\documentclass[letterpaper]{article} 
\usepackage[preprint]{aaai2027} 
\usepackage[hyphens]{url} 
\usepackage{graphicx} 
\def\UrlFont{\rm} 
\usepackage{natbib} 
\usepackage{caption} 
\usepackage{booktabs}
\usepackage{amsfonts}
\usepackage{amsmath}
\usepackage{amssymb}
\usepackage{microtype}

\newcommand{\benchname}{\textsc{Whiteboard}}
\newcommand{\clamp}{\mathrm{clip}_{[0,1]}}
\newcommand{\raw}{\mathrm{raw}}
\newcommand{\pure}{\mathrm{pure}}
\newcommand{\gated}{\mathrm{gated}}
\newcommand{\resid}{\mathrm{resid}}

\newcommand{\atoms}{\mathcal{A}}
\newcommand{\isubs}{\mathcal{I}}
\newcommand{\hsubs}{\mathcal{H}}
\newcommand{\destructive}{\mathcal{H}_{\mathrm{dest}}}

\newcommand{\Iassoc}{I_{\mathrm{assoc}}}

\newcommand{\Hlogic}{H_{\mathrm{logic}}}

\newcommand{\Hctx}{H_{\mathrm{ctx}}}
\newcommand{\Hdetail}{H_{\mathrm{det}}}
\newcommand{\Hfact}{H_{\mathrm{fact}}}

\newcommand{\Hintent}{H_{\mathrm{int}}}

\newcommand{\finding}[2]{\noindent\textbf{Finding #1.}~\emph{#2}}

\newcommand{\techapp}{technical appendix}
\newcommand{\apptaxonomy}{A}      
\newcommand{\appatomaudit}{B}     
\newcommand{\appdecouple}{C}      
\newcommand{\appfullcells}{D}     
\newcommand{\appforest}{E}        
\newcommand{\apppaired}{F}        
\newcommand{\appsensitivity}{G}   
\newcommand{\appradar}{H}         
\newcommand{\appcase}{J}          
\newcommand{\apppanel}{K}         
\newcommand{\apphyperparams}{L}   
\newcommand{\appheldout}{M}       
\newcommand{\appreliability}{R}  

\title{Is Imagination Derived from Hallucination?\\
A Cross-Taxonomy Evaluation of Imagination and\\
Hallucination in Large Language Models}

\author{
Zixuan Tang\textsuperscript{1,}\equalcontrib,\quad
Hongzong Li\textsuperscript{2,3,}\equalcontrib,\quad
Shuxin Zhuang\textsuperscript{1},\quad
Dapeng Wu\textsuperscript{1},\quad
Zi Liang\textsuperscript{4,}\corresponding
}
\affiliations{
\textsuperscript{1}City University of Hong Kong\quad
\textsuperscript{2}Northwestern Polytechnical University\\
\textsuperscript{3}The Hong Kong University of Science and Technology\quad
\textsuperscript{4}The Hong Kong Polytechnic University\\[0.2em]
{\small\texttt{\{zixuatang6-c, shuxin.zhuang\}@my.cityu.edu.hk;\ dapengwu@cityu.edu.hk}}\\
{\small\texttt{lihongzong@nwpu.edu.cn;\ lihongzong@ust.hk;\ zi1415926.liang@connect.polyu.hk}}
}

\begin{document}

\maketitle

\begin{abstract}
Imagination performs as a high-level function of large language models
(LLMs) which determines the potential of how an LLM creates unseen or
creative content. While existing works have built a rich family of
creativity benchmarks for this ability, they only measure how far an
output departs from common answers and never check whether the
departure is licensed by the prompt. Moreover, hallucination, the
closest neighbor of imagination, is always measured in a separate pipeline on different
generations, so the influential claim that imagination and hallucination
stem from the same generative mechanism has never been directly
testable. In this paper, we propose \benchname{}, the first LLM imagination evaluation benchmark. Its design follows the
authoritative cognitive instruments developed to measure human
imagination: seven mechanism-grounded imagination subtypes are adapted
from classic paradigms, then crossed with ten
support-boundary hallucination subtypes and scored jointly on the same
generation. Different from
previous creativity or hallucination
benchmarks, \benchname{} gates every imagination score with an explicit
support check and computes both axes deterministically through an
 auditable atom matrix, with no LLM judge on the primary path.
The full \benchname{} item bank contains 1,660 prompts; on its shared 80-item anchor set, we evaluate 79 state-of-the-art LLMs and validate the instrument against 13,280 human judgments.
Additionally, we further explore
whether imagination derives from the same generative tendency as
hallucination and what key factors shape it. \textbf{Our analysis
indicates a counterintuitive correlation between hallucination and imagination:
Most of the subtype couplings are negative, every one of the anchor items reproduces the negative coupling on its own.}
\end{abstract}

\vspace{0.9em}
\noindent{\large\textbf{Code:}}\\[0.15em]
{\small\color{blue!65!black}\def\UrlFont{\ttfamily}%
\url{https://github.com/erictang666/WHITEBOARD}}

\section{Introduction}
\label{sec:intro}

Imagination is the capacity to produce novel and appropriate content. In
large language models, imagination ranks as a high-level capability. A
model with imagination moves beyond restating memorized text, and
frontier systems increasingly separate from merely competent ones along
this axis~\citep{guilford1967nature,finke1992geneplore}. LLM-generated
research ideas already outscore human ideas on novelty, though they
still trail on feasibility~\citep{si2025llmideas}. An influential
account ties this capability to hallucination: confabulated outputs
reportedly show greater narrativity and semantic coherence than
veridical ones~\citep{sui2024confabulation}. Read at face value, the
account implies an uncomfortable trade-off. A more imaginative model
must be a less trustworthy one. Yet no study has tested the trade-off on
a single output. Any such test must score both axes at once.

\begin{figure}[t]
\centering
\includegraphics[width=\linewidth]{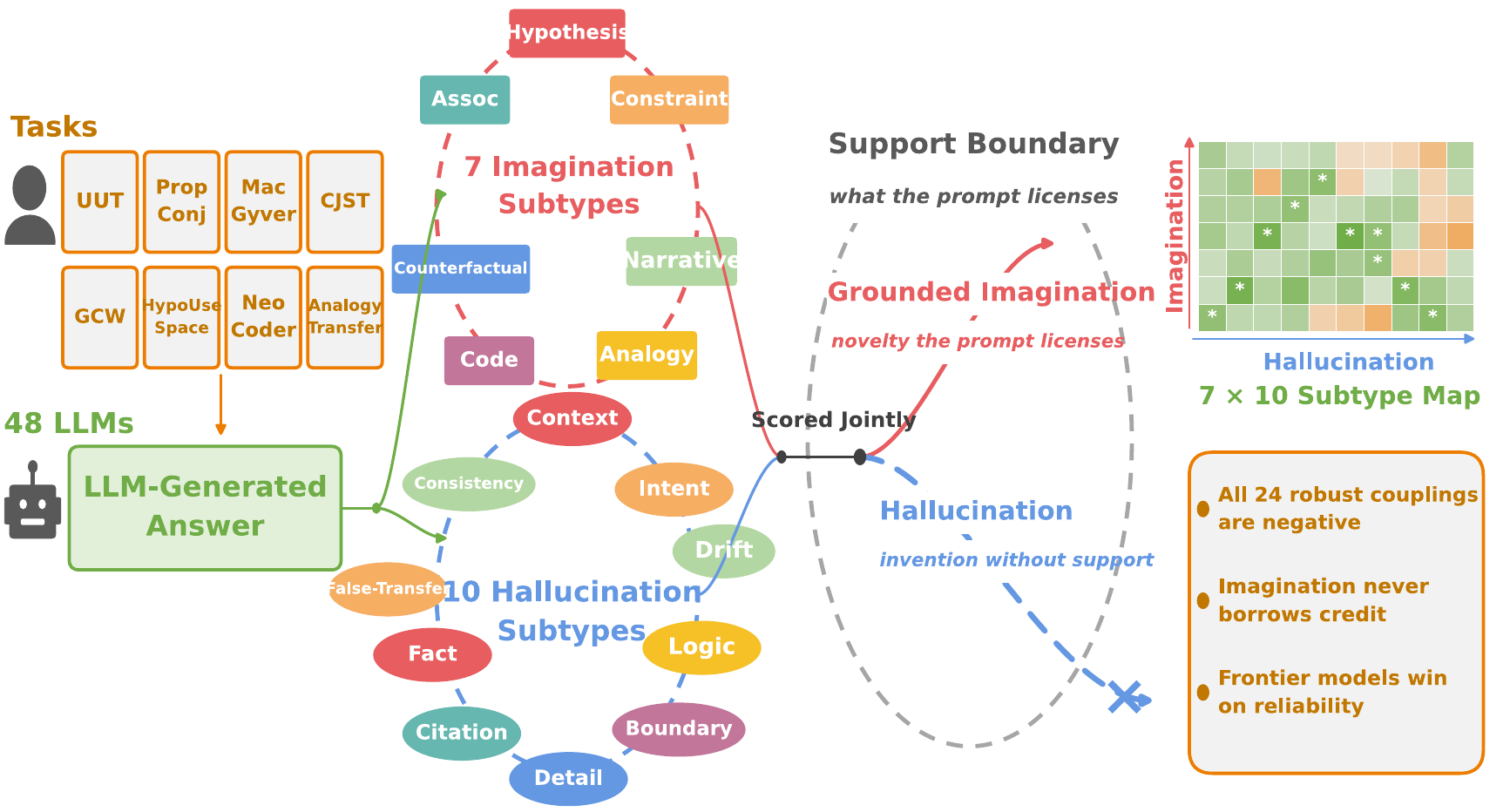}
\caption{\benchname{} reads each response twice, treating novelty
licensed by the prompt as grounded imagination and unsupported invention
as hallucination. Every model answers the identical anchor set, with
one response per item. The eight task families shown at left
provide the scores on both axes. Licensed novelty is assigned to
seven imagination subtypes. Unsupported content is
assigned to ten hallucination subtypes. The deterministic scoring procedure derives
$103$ signed atom signals from task-specific evidence and embedding
similarities; it does not use an LLM judge. }
\label{fig:overview}
\end{figure}

Existing evaluations cannot test whether more imaginative outputs also
carry more hallucinations. Creativity benchmarks measure how far an output
departs from common answers, through word-association norms and embedding
distance~\citep{olson2021naming}, multidimensional
rubrics~\citep{stevenson2022putting}, holistic or LLM-judge
panels~\citep{hou2025creativityprism,ruan2024liveideabench}, or
constraint-driven problem
solving~\citep{tian2024macgyver,lu2024neocoder,atmakuru2024cs4}. None of
these methods separates licensed novelty from unsupported invention. Each
method rewards a rare output without first asking whether the prompt permits
the departure. Two very different outputs therefore earn the same score. A
feasible repair improvised under a tool constraint scores like a fabricated
API call.

Hallucination research, the closest neighbor of imagination, runs in a
separate pipeline on different generations. This line asks a single
question: does an output stay inside a known support set? Answers range
from binary truthfulness labels~\citep{lin2022truthfulqa,li2023halueval}
through atomic-claim verification~\citep{min2023factscore,wei2024longfact}
to span-level annotation~\citep{bao2024faithbench,niu2024ragtruth}. Every
deviation from the reference counts as a defect, even a deviation the
task invites. The two families therefore never score the same
generation, and the shared-mechanism claim has never been directly
testable. Deployment and fine-tuning decisions nonetheless ride on which
view a team takes~\citep{chiang2024chatbotarena,huang2024hallusurvey}.
The panel design adds a third obstacle. Each model answers a different
subset of items, so a score gap between two models confounds model
quality with item difficulty. Three
questions therefore remain open: \textbf{(1)} what correlation
structure the two axes show when read jointly from one generation;
\textbf{(2)} whether models trade hallucination for imagination, and
whether the frontier tier stratifies along either axis; and
\textbf{(3)} whether any hallucination subtype behaves as
\emph{productive} exploration licensed by task semantics.

In this paper, we propose \benchname{}, a benchmark dedicated to imagination
in LLMs. \benchname{} scores seven mechanism-grounded imagination subtypes and
ten support-boundary hallucination subtypes on one generation
(Figure~\ref{fig:overview}). The panel is fully crossed. All $79$ models
answer the same $80$-item anchor set. Every model comparison therefore becomes
a paired test. Unlike earlier creativity and hallucination benchmarks,
\benchname{} applies an explicit support check to every imagination score. Our
design also pairs each imagination subtype with the hallucination the subtype
risks: grounded narrative with fabricated detail, counterfactual extension
with impossible physics, creative code with API invention, and analogical
mapping with false transfer. Each pairing follows the task mechanism. A valid
counterfactual preserves causal edges and avoids causal jumps. Invention
inside a closed support sheet follows the constrained creative cognition of
\citet{finke1992geneplore}, and the same constraint suppresses unsupported
detail. Scoring stays deterministic and embedding-anchored. The primary path
reads a signed atom-audit matrix instead of calling an LLM judge. No prior
benchmark reads both axes from one generation at subtype resolution. Our
design therefore asks directly whether stronger imagination of one form comes
with more or less hallucination in the matched subtype.

We evaluate 79 contemporary instruction-tuned LLMs from 18 providers on the
shared $80$-item anchor set. The resulting $6{,}229$ valid crossed
generations yield the full correlation map. The map is sparse and one-sided.
Of $189$ candidate cells, $23$ survive the triple false-discovery contract,
and $22$ of the survivors are negative. All $55$ items with sufficient score
variation reproduce the negative coupling. Fifty-four items reach individual
significance, and none turns positive (median $\rho{=}{-}0.63$, sign test
$p{=}5.6{\times}10^{-17}$). The coupling survives controls for output length
and for release date. We validate the instrument against $13{,}280$ blind
judgments from two independent annotators on $6{,}640$ anchor outputs. At the
output level, \benchname{} tracks imagination and hallucination ratings at
$\rho{=}0.72$ and $0.69$. At the model level, the same correlations reach
$0.56$ and $0.67$. The annotators lend no differentiated support to the
popular ``productive hallucination'' hypothesis. Across all ten subtypes,
annotators license unsupported content at a nearly uniform rate of
$18.7$--$22.8\%$. These couplings remain correlational rather than causal.
The couplings are nevertheless one-sided, reproducible item by item, and
anchored in human judgment. At subtype resolution, our findings answer the
question in the title. Imagination does not derive from hallucination.
Imagination acts as its antidote.

\paragraph{Contributions.}
\begin{enumerate}
  \item \textbf{\benchname{} scores imagination and hallucination on the
        same model response.} No earlier benchmark reads both axes from one
        output. Our support check gates each imagination subtype and pairs
        it with the hallucination its task invites.
  \item \textbf{Scoring without a judge.} Task families feed raw atoms into
        a signed audit matrix. Scoring stays deterministic and
        embedding-anchored, and no model-as-judge call enters the primary
        path.
  \item \textbf{Removing the item confound.} Panel benchmarks assign each
        model a different item subset, so item difficulty contaminates
        every score gap. \benchname{} removes this confound. All $79$
        models answer an identical anchor set. Each model comparison
        therefore becomes a paired test.
  \item \textbf{We find imagination to be the antidote to hallucination,
        not its byproduct.} The subtype map shows a one-sided negative
        correlation. Every eligible anchor item reproduces the coupling.
\end{enumerate}

\section{Related Work}
\label{sec:related}

\begin{table}[t]
\centering
\includegraphics[width=\linewidth]{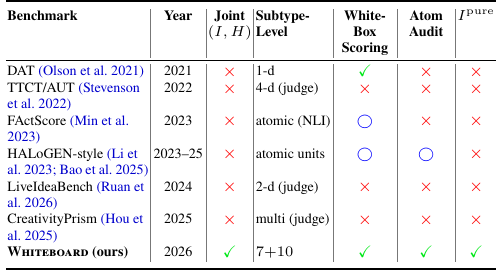}
\caption{Capability comparison with representative baselines. Joint
$(I,H)$ means scoring both axes on the same generation; white-box means
the primary scoring path has no model-as-judge call. The Joint $(I,H)$
column is the precise sense in which \benchname{} is first.}
\label{tab:competitor_features}
\end{table}

\subsection{Single-Axis Benchmarks}
\label{sec:rel_singleaxis}

Existing benchmarks treat creativity and hallucination as separate
evaluation targets. Creativity benchmarks measure how far a response sits
from a reference distribution. The distance takes many forms: common-answer
banks and word norms~\citep{olson2021naming}, associative chains grown from
seed words~\citep{gray2019forward}, TTCT-style rubrics with several raters
per output~\citep{stevenson2022putting}, and holistic or idea-rating panels
scored by an LLM judge~\citep{hou2025creativityprism,ruan2024liveideabench}.
Other designs constrain the task itself, through object reuse, mandatory
story constraints, and denial prompting on
code~\citep{tian2024macgyver,atmakuru2024cs4,lu2024neocoder}. Newer panels
broaden the formats and keep the divergence-from-reference
framing~\citep{alrabeyah2025do}. Hallucination benchmarks run the other way.
Coverage grows from binary truthfulness
labels~\citep{lin2022truthfulqa,li2023halueval} through atomic-claim
verification~\citep{min2023factscore,wei2024longfact} to span-level
annotation~\citep{niu2024ragtruth}, and on to taxonomies of error source and
of intrinsic versus extrinsic
failure~\citep{ravichander2025halogen,bang2025hallulens}. Every deviation
from the reference still counts as a defect~\citep{huang2024hallusurvey}.
Neither line asks whether the local world of the prompt licenses the
divergent move. A feasible improvised repair and a fabricated API call
therefore score alike by construction. \benchname{} keeps the divergence
side of the first line and the support-boundary check of the second, and our
support gate lands both readings on the same output.

\subsection{Joint Imagination--Hallucination Measurement and
Interpretability}
\label{sec:rel_joint}

A growing line asks whether the two axes share generative machinery.
Confabulated outputs carry more narrativity than veridical ones, a pattern
suggesting partly shared mechanisms~\citep{sui2024confabulation}. LLM
assistance can raise creativity during assisted tasks and hinder later
unassisted performance~\citep{kumar2024human}. Creativity metrics can
disagree across domains and shift under minor prompt
variations~\citep{lu2025rethinking}. Other results pull the other way.
Inter- and intra-model homogenization complicates distance-based creativity
scores~\citep{jiang2025artificialhivemind}. Anti-hallucination interventions
are judged by how far they move creativity scores on code and story
tasks~\citep{lu2024neocoder}. LLM ideas outrank human ideas on novelty and
trail on feasibility~\citep{si2025llmideas}. Psychometric work had already
framed useful imagination as the move respecting constraints, not the move
ignoring them~\citep{finke1992geneplore}. Three gaps persist across this
line: the same panel rarely receives both scores on one output, the raw
signal flow rarely becomes visible, and the productive-versus-destructive
label rests on stipulation rather than
test~\citep{huang2024hallusurvey}. Prior work treats the shared-mechanism
question as a stance. \benchname{} treats the same question as a falsifiable
proposition and adjudicates it empirically.

\section{The \benchname{} Benchmark}
\label{sec:benchmark}

\subsection{Design Goals}
\label{sec:design_goals}

We fixed four criteria before building \benchname{}. First, one
test-case generation must carry both the imagination score and the
hallucination score. Only such a joint reading, in our view, keeps the
boundary between licensed novelty and unsupported invention inside the
measurement. Second, the two axis-level scores are not enough: each axis
must break into mechanism-grounded subtypes, so a coupling can be
localized rather than averaged away. Third, the primary scoring path
must contain no model-as-judge call, and every atom-to-subtype
contribution must be recorded with a sign. Finally, we must preregister
the aggregation weights, gates, and residualization coefficients, and
every conclusion must survive partial controls for shared formula atoms
and for capability tier.

\subsection{Joint Cross-Taxonomy}
\label{sec:taxonomy}

We survey both literatures, extract candidate subtypes, and consolidate
them under three constraints. Every subtype must be computable from a
deterministic, auditable signal. Every subtype must be carried by at
least one task family. Facets serve as second-level explanatory
variables for the audit, never as new ranking weights.
Table~\ref{tab:taxonomies} lists the seven imagination and ten
hallucination subtypes, with definitions and primary task carriers.
We give the full facet tables in Appendix~\apptaxonomy{} of the
separately submitted \techapp{}. We settle the destructive-versus-productive status of
each hallucination subtype empirically, never by definitional fiat.

\begin{table}[t]
\centering
\includegraphics[width=\linewidth]{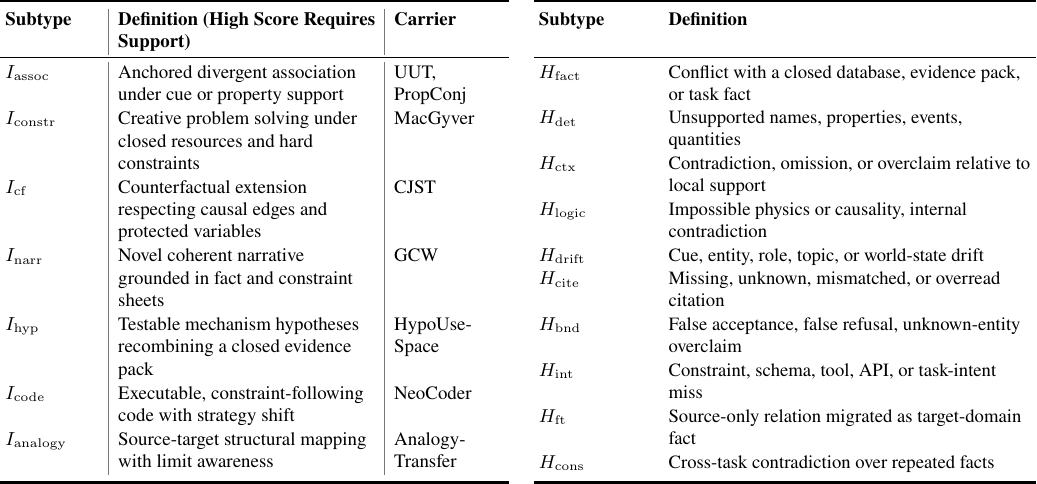}
\caption{Taxonomy subtypes. Left: imagination subtypes ($\isubs$), each
combining a divergence component with a support gate. Right: hallucination
subtypes ($\hsubs$), each computed from a distinct support-boundary check.}
\label{tab:taxonomies}
\end{table}

\subsection{Task Families and Scoring Pipeline}
\label{sec:pipeline}

The pipeline maps a model's generations into the two subtype vectors
$\{I_{m,a}\}$ and $\{H_{m,b}\}$. The primary path stays deterministic,
with no model-as-judge call. A subset of the atoms uses the
\texttt{all-mpnet-base-v2} Sentence-BERT encoder as an embedding anchor
\citep{reimers2019sbert,sentence_transformers2025allmpnet}. Every
atom-to-subtype contribution enters the record as a signed entry, so an
auditor can retrace each score.
Figure~\ref{fig:overview} traces the full flow from task families
through the signed atom matrix to the correlation map.

\paragraph{Task Families.}
The suite defines nine task families, and eight of them carry an
imagination subtype. \textsc{ClosedWorldFact} serves hallucination
calibration alone, contributes no imagination signal, and stays out of
the anchor set used here. Each model answers each anchor item once.
Temperatures are fixed per family: $0.85$ for the creative families,
$0.70$ for MacGyver, and $0.55$ for Forward Flow. Every family also
carries a task-specific JSON-only output contract and a preregistered
token cap. We record parse validity, finish reason, truncation, and
schema coverage, so a missing output reduces eligibility instead of
disappearing silently. Each family carries exactly one imagination
subtype together with the two to four hallucination subtypes its prompts
put at risk, as named in Table~\ref{tab:taxonomies}. DAT, CDAT, and
Forward Flow remain auxiliary association probes, and neither
$I^{\pure}$ nor the correlation map admits them.
Table~\ref{tab:dataset_stats} collects the benchmark's headline
statistics.

\begin{table}[t]
\centering
\includegraphics[width=\columnwidth]{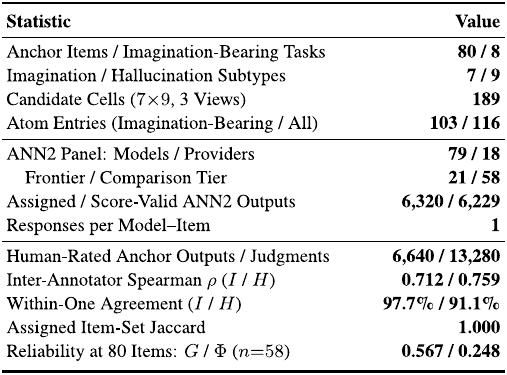}
\caption{Benchmark and evaluation statistics at a glance. Every model is
assigned the identical item set, so item difficulty cannot be confounded
with model identity.}
\label{tab:dataset_stats}
\end{table}

\paragraph{Atoms and Audit Matrix.}
Each scorer emits raw atoms: rarity against a curated common-answer bank,
affordance support, unavailable-tool rate, causal-edge support,
forbidden-update rate, claim-support precision, citation mismatch rate,
hidden-test pass rate, gold-mapping coverage, false-transfer rate, and
others. Lexical atoms draw on the SWOW-EN2018 association norms \citep{dedeyne2019smallworld},
Word Norms 2 \citep{buchanan2019english}, and WordNet 3.0 \citep{miller1995wordnet}.
We record the exact snapshots and access terms in Appendix~\apphyperparams{}.
Across the imagination-bearing families, the pipeline exposes $103$ of its
$116$ atom entries (Appendix~\appatomaudit{}). The audit matrix
$M\in\{-1,0,+1\}^{|\atoms|\times(|\isubs|+|\hsubs|)}$ records the role of
each atom in each subtype: positive contribution, penalty, or no use.
The scoring code generates $M$ automatically, and the shared-atom partial
control below runs on $M$.

\paragraph{Three Views.}
We report every subtype score in three preregistered views. Each view
guards against a different way a single aggregate could mislead. For
model $m$ and subtype $a$, let $x_{m,a}$ collect the raw atoms, and let
$A(\cdot\,;w)$ denote the fixed aggregation operator with preregistered
weights $w$:
\begin{equation}
\begin{aligned}
I^{\raw}_{m,a} &{=} A(x_{m,a};w^{\raw}),\cr
I^{\gated}_{m,a} &{=} I^{\raw}_{m,a}\,g_{m,a},\cr
I^{\resid}_{m,a} &{=} \clamp\!\bigl(I^{\gated}_{m,a}\cr
&\qquad {}{-}\beta_{f}\,\bar{H}^{\raw}_{m,f}\bigr).
\end{aligned}
\end{equation}
The raw view aggregates the atoms as they stand. The gated view
multiplies in $g_{m,a}\in[0,1]$, the carrier family's own support
gate: appropriateness in UUT, constraint compliance in MacGyver and
GCW, premise consistency in CJST. Divergence therefore earns credit
only after the licensing checks pass. The residual view removes
mechanical cross-axis coupling. Here $\bar{H}^{\raw}_{m,f}$ is the mean
raw score of the destructive hallucination subtypes carried by the
same family $f$, and $\beta_{f}$ is the family's preregistered leakage
coefficient, written $\beta^{IH}_{f}$ when the direction matters.
We clip the difference back to $[0,1]$. The H side follows
symmetrically with $\beta^{HI}_{f}$. A conclusion appearing in
only one view counts as a leakage symptom, not as a finding,
and the partial controls below remove what the views cannot.

\paragraph{Correlation Analysis.}
We build the map cell by cell. For a view $v$ and a pair $(a,b)$, the
cell statistic is the Spearman rank correlation $\rho^{v}_{a,b}$
between the panel's score vectors $\{I^{v}_{m,a}\}_{m}$ and
$\{H^{v}_{m,b}\}_{m}$. Each cell carries a $95\%$ bootstrap
confidence interval ($B{=}2{,}000$ resamples, seeded from a stable
per-cell hash), a two-sided $p$-value, and BH-FDR $q$-values. We compute
the $q$-values within each view and globally over all
$7{\times}9{\times}3{=}189$ cells. The \emph{consistency} subtype stays
out of the map, as a zero-weight diagnostic carried by a single
auxiliary probe. Two partial estimates
then ask whether a surviving correlation is an artifact. The
shared-atom partial $\rho^{\mathrm{a}}_{a,b}$ holds the atoms feeding
both scores fixed: we rank-transform $I_a$, $H_b$, and the shared atoms
$S_{a,b}{=}\{\alpha:M[\alpha,I_a]\,M[\alpha,H_b]\neq 0\}$, regress
both score ranks on the atom ranks, and correlate the residuals. An
empty $S_{a,b}$ returns $\rho^{v}_{a,b}$ unchanged. The
capability partial $\rho^{\mathrm{c}}_{a,b}$ asks whether the
correlation is only a shadow of general capability, and conditions the
same way on the proxy $C_m$ (a frozen external Arena score where available,
otherwise the mean of $\{I_{m,a'}\}_{a'\neq a}$). We distinguish the official
LMArena sources from the third-party ones in Appendix~\apphyperparams{}. Writing
$q^{\mathrm{v}}_{a,b}$, $q^{\mathrm{a}}_{a,b}$, and
$q^{\mathrm{c}}_{a,b}$ for the $q$-values of the view test and of the
two partials, we promote a cell to a main claim only when
$\max\bigl(q^{\mathrm{v}}_{a,b},q^{\mathrm{a}}_{a,b},q^{\mathrm{c}}_{a,b}\bigr)\leq 0.05$.

\paragraph{Purified Imagination.}
The purified total serves one purpose: no model may buy imagination
credit with fabrication. We call the coupling between an
imagination subtype $a$ and a hallucination subtype $b$ \emph{robustly
positive} when both partial estimates find the coupling,
\begin{equation}
\tilde{\rho}_{a,b}=
\left\{
\begin{aligned}
&\min\bigl(\rho^{\mathrm{a}}_{a,b},\rho^{\mathrm{c}}_{a,b}\bigr),\cr
&\quad \text{if both are positive and}\cr
&\quad \max\bigl(q^{\mathrm{a}}_{a,b},q^{\mathrm{c}}_{a,b}\bigr)\leq 0.05,\cr
&0,\quad \text{otherwise.}
\end{aligned}
\right.
\label{eq:robust}
\end{equation}
With $w_a{=}1/|\isubs|$ and $\destructive$ the destructive set, we
define the purification coefficient and the
purified total as
\begin{equation}
\pi_a=1-\max_{b\in\destructive}\tilde{\rho}_{a,b},
\qquad
I^{\pure}_{m}=\frac{\sum_{a}\pi_a w_a I_{m,a}}{\sum_{a}\pi_a w_a},
\label{eq:Ipure}
\end{equation}
both computed on the raw-view partial estimates. A subtype never
borrowing against destructive hallucination keeps $\pi_a{=}1$ and
its full weight. A subtype whose score rises with some destructive
subtype, even after both controls, loses exactly the strength of that
coupling. On the current panel, every $\pi_a$ equals $1$. We
preregistered the device nonetheless, so
any future model inflating creativity through fabricated affordances
or facts loses weight automatically.

\section{Experiments}
\label{sec:results}

We first describe the evaluation panel and the shared anchor design. We
then follow the three questions posed in the Introduction: the joint
correlation structure, who trades imagination for reliability, and
whether any hallucination is productive. Human validity comes last. We
report the reliability boundaries of the instrument in
Appendix~\appreliability{}. Six audited outputs, subtype profiles by
tier, and the forest plot of the decisive cells appear in
Appendices~\appcase{}, \appradar{}, and~\appforest{}.

\subsection{Empirical Setup}
\label{sec:setup}

The main panel evaluates $79$ contemporary instruction-tuned LLMs from
$18$ providers. Release recency and product position split the panel
into $21$ frontier-tier and $58$ comparison-tier systems. We list the
model families and their release references in Appendix~\apppanel{}.
Four providers contribute most of the panel, so we report provider
leave-one-out robustness for all $18$ (Appendix~\appreliability{}).
Every model answers the same $80$ anchor items once. The design
therefore defines $6{,}320$ model-item cells. Of these cells, $6{,}229$
pass the task's parse contract, and $91$ stay recorded as missing rather
than refilled. Fifty-eight models carry a scored output for every item.
Generation disables reasoning modes and uses fixed output counts and
task-specific JSON contracts. We fixed the scorer hyperparameters before
the analyses reported here, and only the first annotator's ratings
informed that choice. The second annotator stays held out. Runtime
scoring never branches on model identity, release date, or provider.
Table~\ref{tab:panel_roster} lists the panel, and we give the
hyperparameters in Appendix~\apphyperparams{}.

\begin{table*}[t]
\centering
\includegraphics[width=\textwidth]{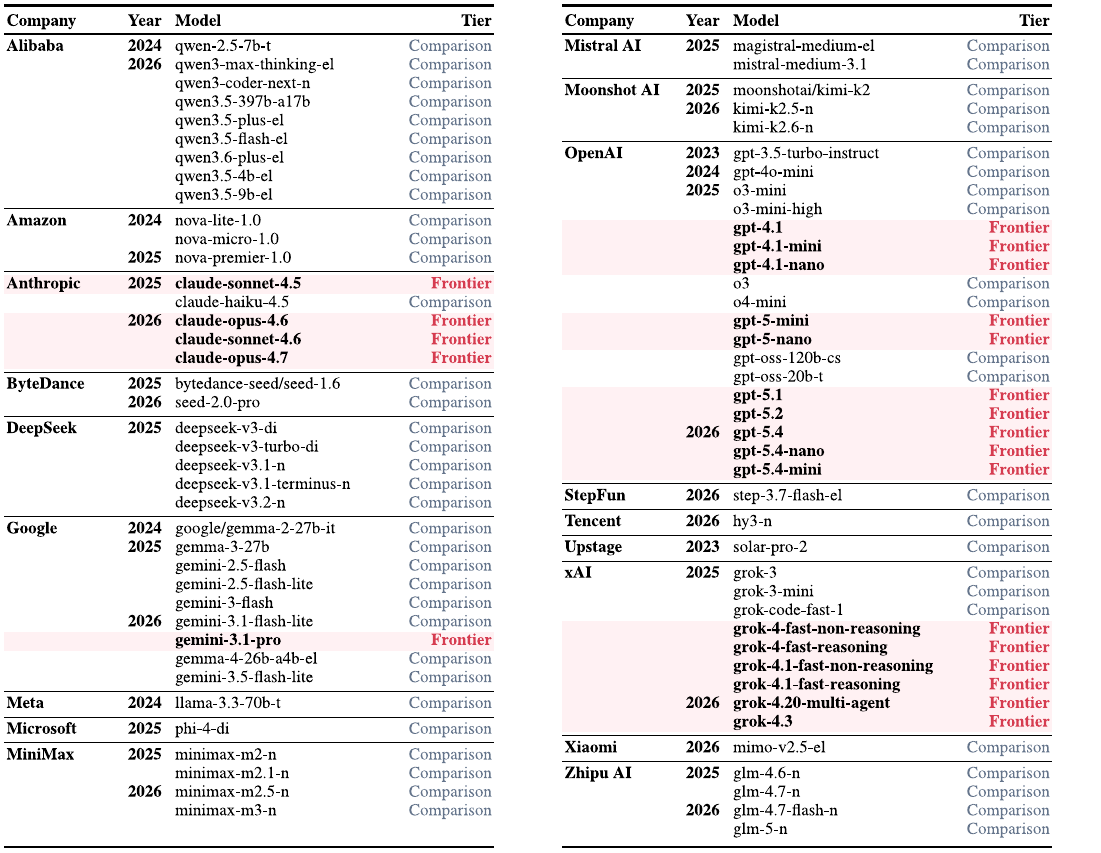}
\caption{The $79$-model panel by provider and release year. Tier
assignment reads model metadata only and never reads scores.}
\label{tab:panel_roster}
\end{table*}

\subsection{The Anchor Set and Paired Discriminative Power}
\label{sec:results_anchor}

Question (1) stands or falls before any correlation is computed. The
panel must separate model differences from item difficulty, so we begin
by measuring what the anchor design buys. Every model receives the
identical anchor set (Table~\ref{tab:dataset_stats}). We compare the
item-set overlap of that design against our earlier prompt-collection
design. On the models with complete coverage, we then test all $1{,}653$
model pairs for a difference in imagination, twice over identical
scores: once paired across shared items, once unpaired, both at BH-FDR
$q{\leq}0.05$. Assigned overlap reaches $1.000$ in Jaccard terms, and
$0.972$ once score-invalid outputs are dropped. The earlier collection
reaches $0.003$, with $89\%$ of items answered by exactly one model. The
paired test resolves $54$ pairs ($3.3\%$), and the unpaired test
resolves none (Appendix~\apppaired{}). The variance explains the
asymmetry: items account for $74.5\%$ of the per-item score variance,
models for $0.41\%$ (Appendix~\appreliability{}). Difficulty therefore
buries the between-model signal unless the item stays fixed, and pairing
is what removes the difficulty term. The anchor design is no
convenience. The design is the precondition for the rest of this
section, and the couplings below are statements about models rather than
about which prompts a model happened to receive.

\subsection{The Subtype Correlation Map}
\label{sec:results_map}

\begin{figure*}[t]
\centering
\includegraphics[width=\textwidth]{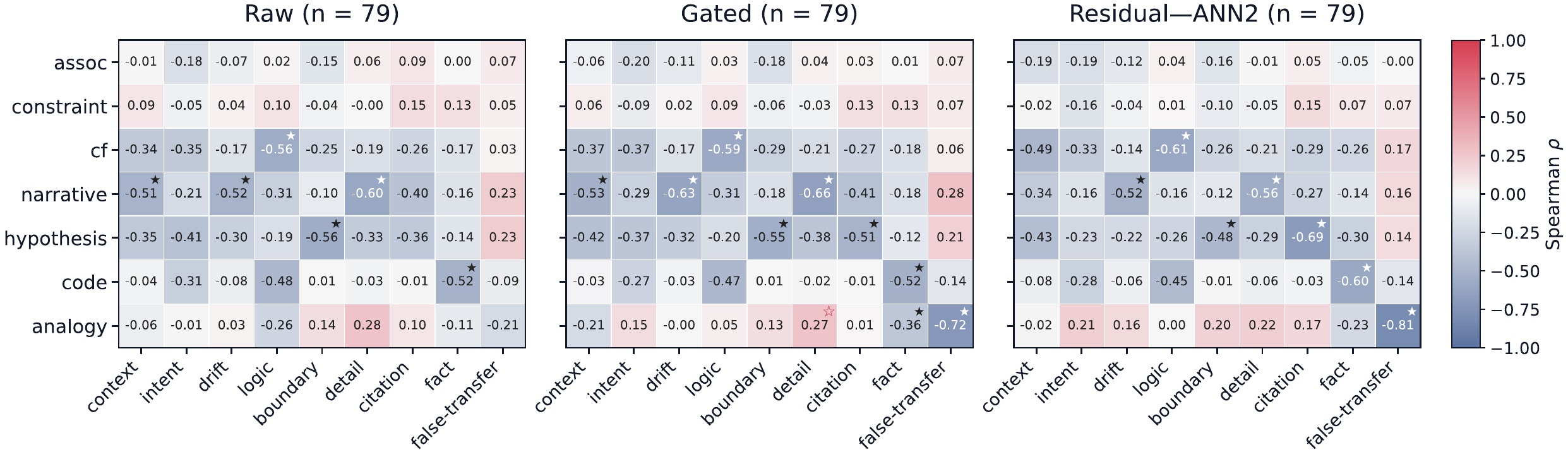}
\caption{Subtype correlation map ($7{\times}9$, three views: raw,
gated, residual). Cell text is Spearman $\rho$ over $79$ models; filled
stars mark cells passing the triple-FDR contract (view-, atom-, and
capability-partial BH-FDR at $q{\leq}0.05$) with a negative
coefficient, and the open star marks the single cell that passes with a
positive one. First, only the starred pairs expose a robust coupling
between an imagination subtype and its partner hallucination subtype;
every unstarred cell is statistically indistinguishable from zero under
the triple correction. Second, among the couplings that do emerge, the
direction is overwhelmingly negative (cool colors): $22$ of the $23$
starred cells lie below zero.}
\label{fig:heatmap_main}
\end{figure*}

We next read the sign structure of the two axes directly, at the
resolution where the mechanism claim lives. The map holds $189$
candidate cells over the $79$-model panel ($7$ imagination subtypes
$\times$ $9$ hallucination subtypes $\times$ $3$ views). A cell earns
promotion only after clearing BH-FDR at $q{\leq}0.05$ in its own view
and in both partial controls. Figure~\ref{fig:heatmap_main} shows the
three views. Table~\ref{tab:correlation_cells} lists the strongest cells
together with the distribution behind them. Twenty-three cells clear the
contract, and $22$ of them are negative ($6$ raw, $9$ gated, $7$
residual). The bulk of the map points the same way: $68$--$76\%$ of
cells are negative in each view, and the per-view median $\rho$ falls
between $-0.10$ and $-0.14$.

\finding{1}{Of $189$ candidate cells, $23$ survive the triple-FDR
contract and $22$ of them are negative: doing an imagination subtype
well predicts fewer of its partner hallucinations, not more.}

The decisive cells pair each imagination subtype with the failure its
own discipline forbids: analogical mapping against false transfer
($-0.81$ residual), hypothesis generation against citation mismatch
($-0.69$), grounded narrative against unsupported detail ($-0.66$
gated) and entity drift ($-0.63$), counterfactual extension against
logic violation ($-0.61$), and creative code against fabricated APIs
($-0.60$). Every one of these moves is a single operation read from two
sides: separating source-only relations from target facts, sourcing a
claim, staying inside a fact sheet, respecting causal edges, and
importing only what exists. One cell passes in the positive direction,
analogical mapping against unsupported detail in the gated view
($\rho{=}{+}0.27$, $q{=}0.042$). This cell rests on no such mechanism
and sits at the correction threshold, so we report a threshold artifact
rather than evidence for a productive hallucination subtype. We test the
productive-hallucination hypothesis on human judgments instead. The
answer to question (1) is therefore one-sided. Wherever a coupling
between the two axes can be established at all, the two axes almost
always move in opposite directions.

\begin{table}[t]
\centering
\includegraphics[width=\linewidth]{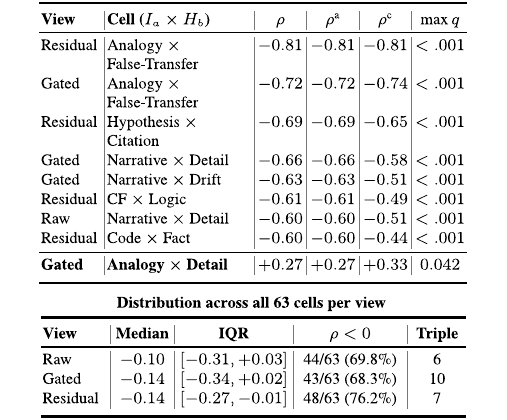}
\caption{Top: the strongest cells passing the triple-FDR contract,
selected from the $23$ listed in full in Appendix~\appfullcells{}; the
set-off row is the only cell that passes in the positive direction.
Bottom: the distribution of all $63$ cells in each view, which shows
that the decisive cells are the tail of an already negative mass rather
than isolated extremes. The atom partial leaves every reported $\rho$
nearly unchanged, because the v3 schema decouples the cross-axis atom
sharing found in the scorer audit (Appendix~\appdecouple{}); the
capability partial reduces some couplings but preserves significance
throughout.}
\label{tab:correlation_cells}
\end{table}

\subsection{One Coupling or Fifty-Five?}
\label{sec:results_peritem}

\begin{figure}[t]
\centering
\includegraphics[width=\linewidth]{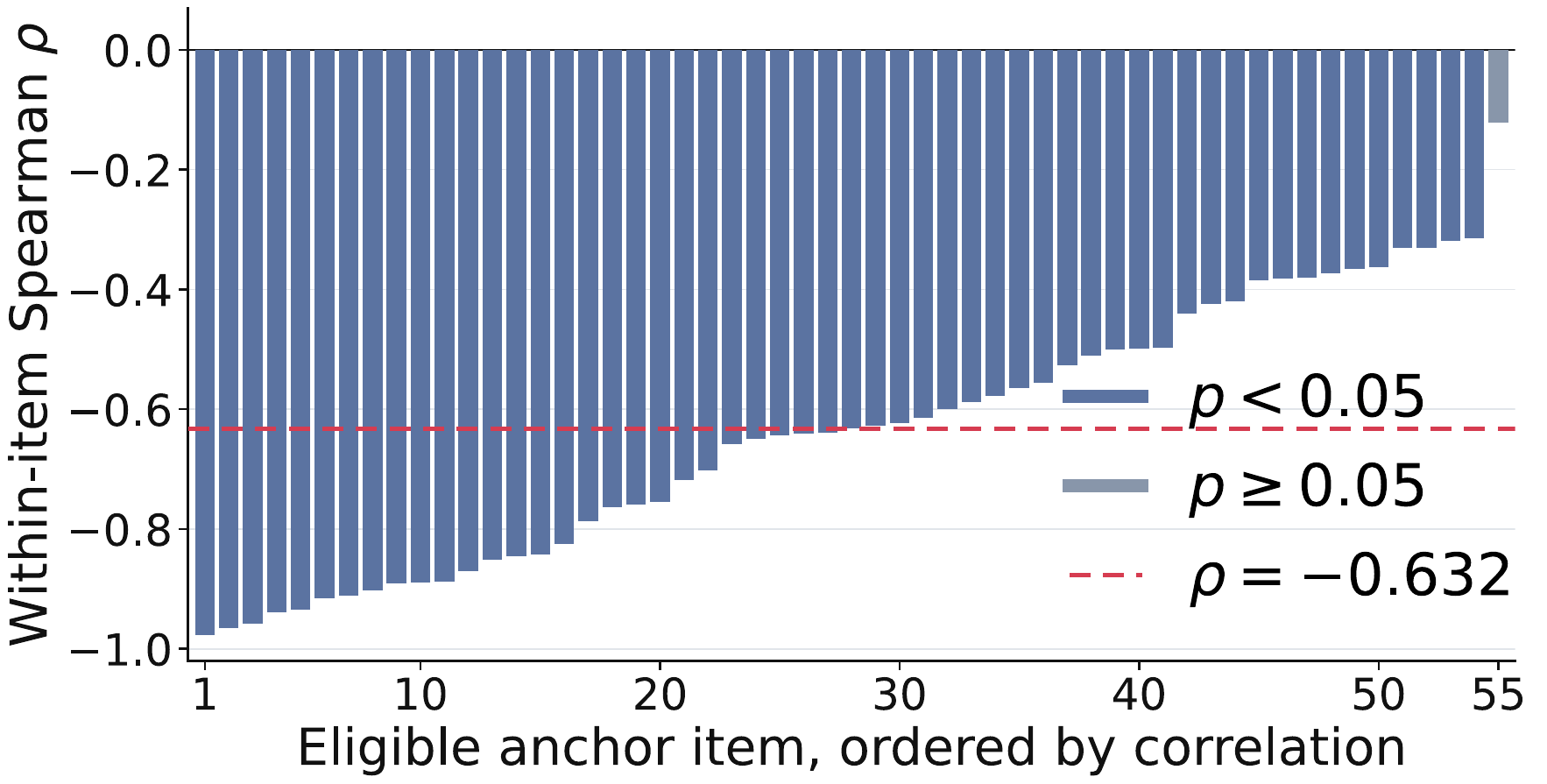}
\caption{Per-item replication. Each bar is the within-item Spearman
correlation between imagination and destructive hallucination across
the models that answered that anchor item; blue bars are individually
significant at $p{<}0.05$. All $55$ eligible items are negative and none
is positive, so the panel-level coupling reproduces item by item rather
than emerging from aggregation.}
\label{fig:per_item}
\end{figure}

A panel-level correlation over $79$ models is a single measurement.
Aggregation can produce such a measurement as easily as the behavior the
measurement should summarize, so we take the panel apart and look for
the coupling again. Figure~\ref{fig:per_item} reports $55$ eligible item
correlations. All are negative, $54$ reach significance at $p{<}0.05$,
and the median is $\rho{=}{-}0.63$. Holding the prompt, the difficulty,
and the license fixed rules out item aggregation. Fifty-five item-level
couplings therefore agree: a model inventing more inside the license
invents less outside it.

\subsection{What Else Could Produce the Coupling?}
\label{sec:results_confound}

A negative association between two axes read from the same generation
invites a third-variable explanation, so we test the three candidates a
reviewer can name from the data alone. Controls for output length and
for release date leave the aggregate coupling near $\rho{=}{-}0.414$.
The capability proxy reduces the coupling to $-0.093$. The proxy is
itself an imagination-level measure, and all $22$ negative subtype cells
already pass the same capability control. Deterministic decoding
strengthens the coupling rather than weakening it
(Appendix~\appsensitivity{}). We therefore keep the aggregate value as a
summary only, and we rest question (1) on the subtype map and its
per-item replication.

\subsection{Who Trades Imagination for Reliability?}
\label{sec:results_purified}

\finding{2}{No imagination subtype borrows credit from destructive
hallucination (all purification coefficients equal $1$), but the
frontier tier separates from the field on neither axis: the trade-off
narrative has no tier-level counterpart on this panel.}

\begin{figure}[t]
\centering
\includegraphics[width=\columnwidth]{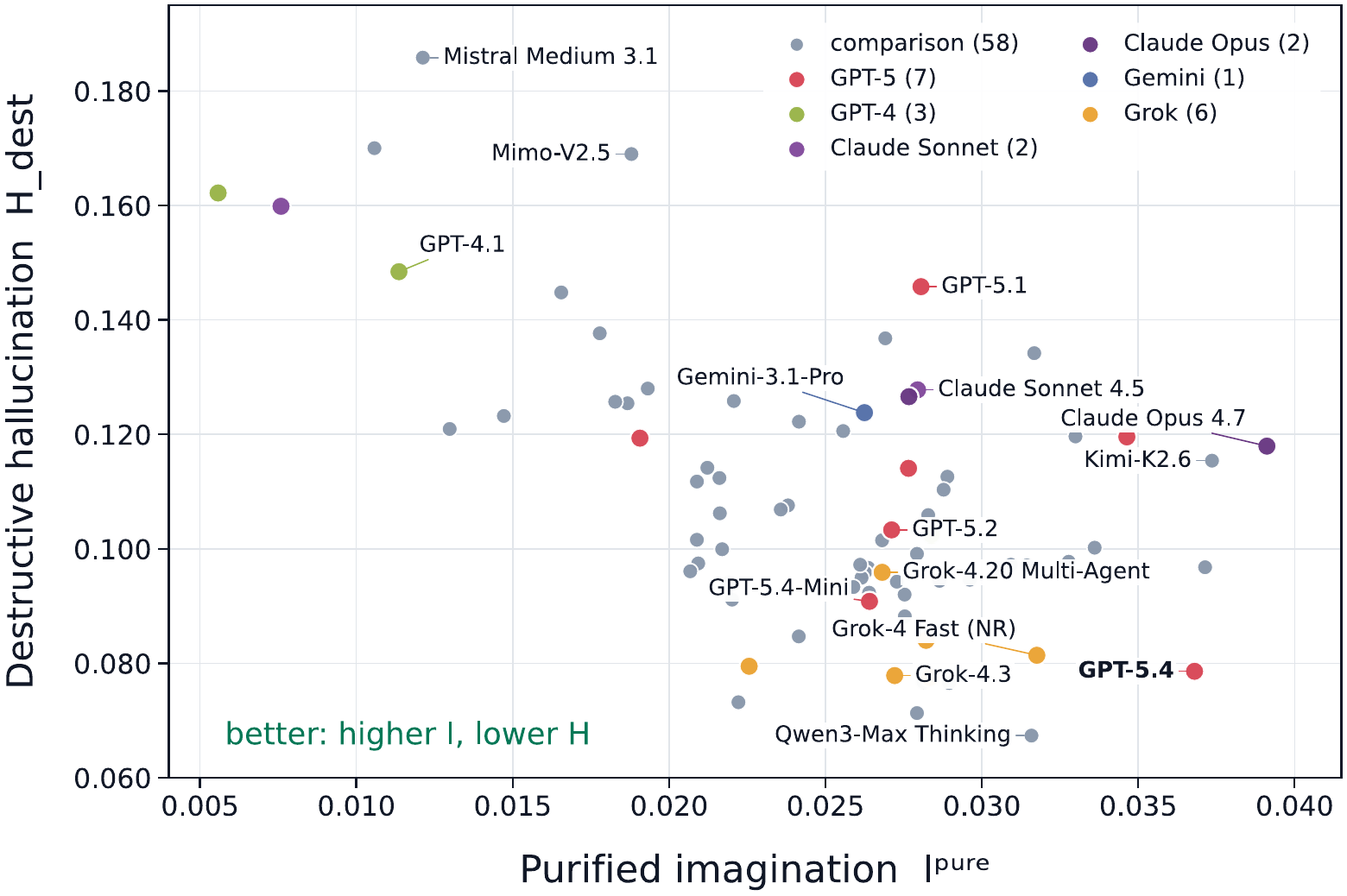}
\caption{The panel on the imagination--hallucination plane; better is
bottom-right. Named points are the flagship families, gray the
comparison tier. The tiers do not separate: the named points are
dispersed through the panel rather than displaced downward, and the
gradient that remains is the coupling of Finding 1, not a tier effect.}
\label{fig:dual_axis}
\end{figure}

Question (2) asks whether the negative coupling shows up as a division
between model tiers. The intuitive reading of the map predicts exactly
such a division: stronger models buy reliability with imagination, or
the reverse. We feed Table~\ref{tab:correlation_cells} into
Equation~\eqref{eq:Ipure} and compare the tiers under seven score-blind
rules. All purification coefficients equal $1$, so the purified total
and the raw total coincide. The tiers separate on neither axis
(Figure~\ref{fig:dual_axis}). The stricter flagship result also fails
Holm correction. Finding 1 is therefore a within-model regularity, and
tiering averages the regularity away.

\subsection{Is Any Hallucination Productive?}
\label{sec:results_productive}

\finding{3}{Across $79$ models and $13{,}280$ blind human judgments, no
hallucination subtype is distinctively licensed: annotators accept
unsupported content at a near-uniform rate whatever kind of unsupported
content it is.}

The taxonomy leaves one question open: is each hallucination subtype
uniformly destructive, or can task semantics license it (e.g.\ fabricated
detail in fiction, source-target transfer in metaphor)? Question (3)
asks exactly that. We test two things: positive panel coupling under
Equation~\eqref{eq:robust}, and blind human labels of licensed content.
No raw-view cell is robustly positive. Annotators license about one fifth
of unsupported content, and the rate stays between $18.7\%$ and $22.8\%$
across subtypes (Fisher $p{=}0.22$). Confabulation may carry positive
value~\citep{sui2024confabulation}. Our data, however, single out no
hallucination subtype as distinctively licensed.

\subsection{Validity: Do Humans See the Same Thing?}
\label{sec:results_validity}

\begin{figure}[!t]
\centering
\includegraphics[width=\linewidth]{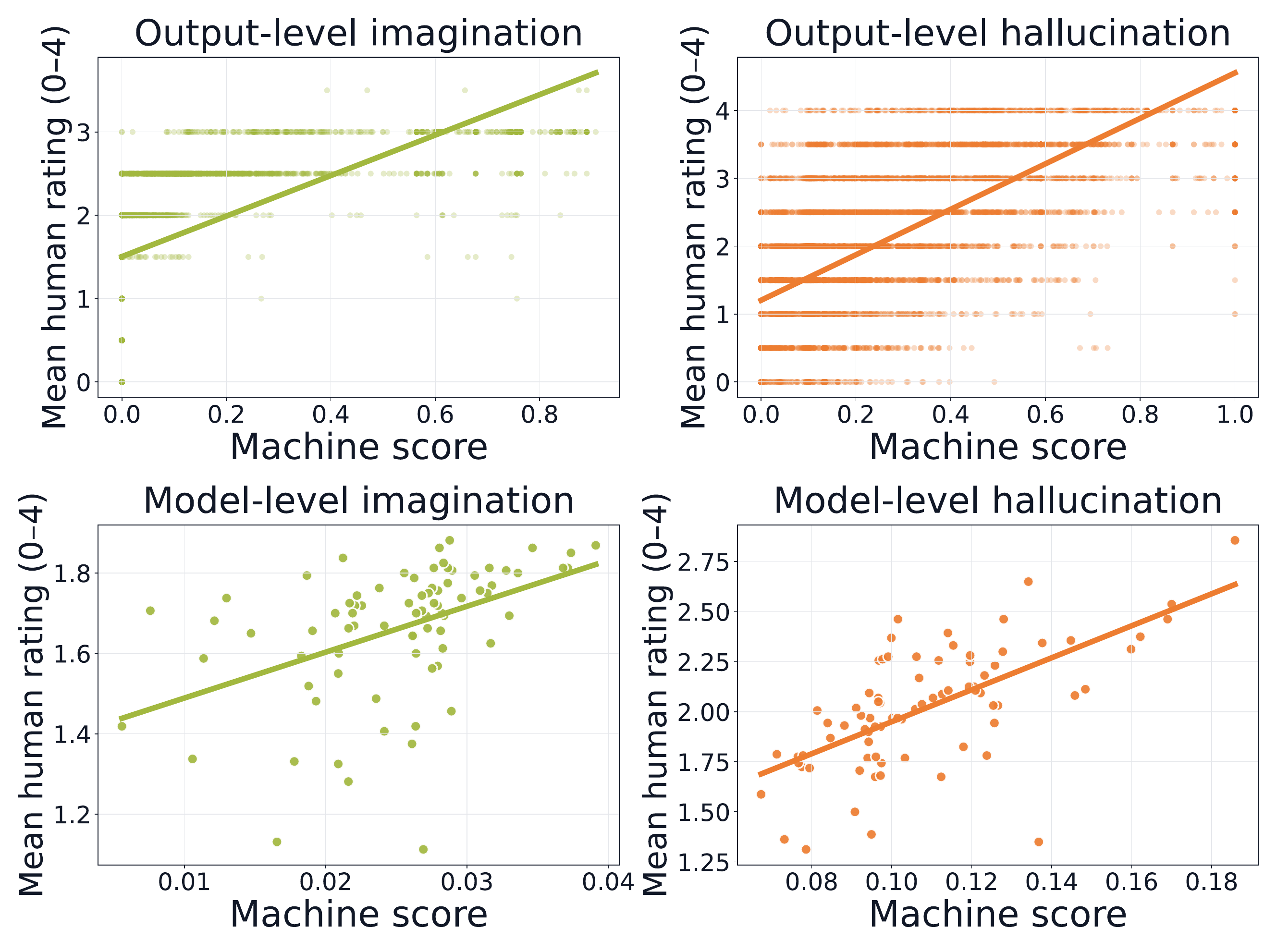}
\caption{Human validity. Top: every valid scored anchor output
($n{=}6{,}443$), machine score against the mean of the two blind
annotators' $0$--$4$ ratings. Bottom: the same comparison aggregated to
the $79$ models, with bootstrap intervals. The footer reports
inter-annotator agreement over all $6{,}640$ annotated outputs. The
instrument tracks human ratings on both axes and at both resolutions.}
\label{fig:human_validity}
\end{figure}

A deterministic scorer produced everything above, so one question
decides the instrument: do the two axes track what people see in the
same outputs? Two blind annotators supplied $13{,}280$ judgments, and
only the second annotator stayed held out from score selection
(Appendix~\appheldout{}). \benchname{} tracks the
held-out annotator at $0.71$ and $0.63$ per output
(Figure~\ref{fig:human_validity}). The human axes, however, give a
coupling of $-0.14$ against the machine score's $-0.41$. Global human
ratings therefore support validity on both axes, but the same ratings
cannot confirm the strength of a subtype-level coupling.

\section{Conclusion}
\label{sec:conclusion}

In this paper we asked whether imagination in large language models
derives from the same generative tendency as hallucination. Existing
evaluations cannot settle the question. The two axes are scored in
separate pipelines on different generations, and a panel where each
model answers different items cannot separate model ability from item
difficulty. \benchname{} reads both axes from one generation over a
shared anchor set. Our support check gates each of the seven imagination
subtypes and pairs each subtype with the hallucination the subtype
risks. A signed atom matrix computes every score, and no LLM judge
enters the path. Across $79$ models and $6{,}229$ crossed generations,
$22$ of the $23$ couplings surviving triple false-discovery control are
negative. Every eligible anchor item reproduces the negative coupling on
its own. Human raters license no hallucination subtype distinctively.
The instrument tracks $13{,}280$ blind judgments at $\rho{=}0.72$ and
$0.69$ per output. At subtype resolution, then, imagination does not
derive from hallucination. Imagination stands against it. We leave the
preregistered purified total and the public audit matrix behind as a
standing test, ready to demote any future model inflating creativity
through fabrication.

\FloatBarrier

\bibliography{refs_arxiv}

\clearpage
\renewcommand{\topfraction}{0.92}
\renewcommand{\bottomfraction}{0.60}
\renewcommand{\textfraction}{0.07}
\renewcommand{\floatpagefraction}{0.75}
\renewcommand{\dbltopfraction}{0.92}
\renewcommand{\dblfloatpagefraction}{0.75}
\setcounter{topnumber}{3}
\setcounter{dbltopnumber}{3}
\setcounter{bottomnumber}{2}
\setcounter{totalnumber}{5}
\emergencystretch=1em
\raggedbottom
\setcounter{secnumdepth}{2}
\makeatletter
\@addtoreset{figure}{section}
\@addtoreset{table}{section}
\makeatother
\renewcommand{\thefigure}{\thesection\arabic{figure}}
\renewcommand{\thetable}{\thesection\arabic{table}}
\newcommand{\mainpaper}{main paper}
\newcommand{\maintabcorrelation}{5}
\newcommand{\mainfigheatmap}{2}
\newcommand{\mainsecpurified}{Who Trades Imagination for Reliability?}
\newcommand{\mainsecanchor}{The Anchor Set and Paired Discriminative Power}
\appendix

\section{Full Taxonomy and Facets}
\label{app:taxonomy}

We will reproduce the full facet tables in the camera-ready, and the
released taxonomy file already carries them. The tables cover all seven
imagination and ten hallucination subtypes, each with its
white-box-signal list. Each subtype has between three and five facets,
and each facet has at least one white-box-signal entry.

\section{Atom Audit Matrix}
\label{app:atom_audit}

\begin{figure*}[t]
\centering
\includegraphics[width=\linewidth]{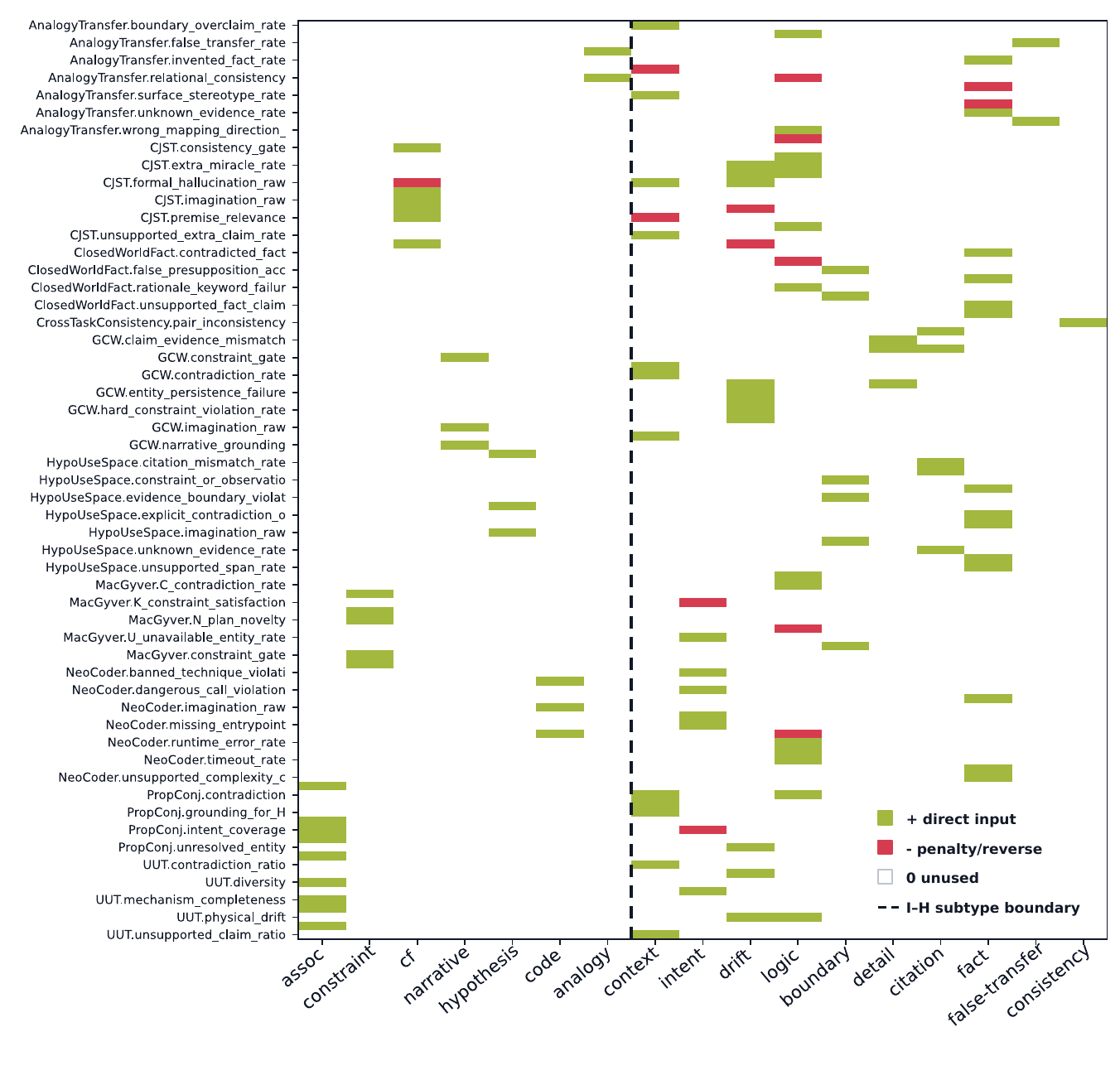}
\caption{Signed atom-to-subtype dependency matrix. Green: positive
contributions; red: penalty signals; white: unused. The block structure
is the substrate for the shared-atom partial Spearman control.}
\label{fig:atom_audit_matrix}
\end{figure*}

The audit matrix $M\in\{-1,0,+1\}^{116\times 17}$ records the signed
contribution of each raw atom entry to each of the $17$
subtype scores. Of the $116$ atoms, $103$ belong to the eight
imagination-bearing families. The remaining $13$ belong to the
hallucination-calibration family and to the zero-weight consistency
probe. Figure~\ref{fig:atom_audit_matrix} visualizes the
sparsity pattern and the separation between I-side support atoms and
H-side penalty atoms. We release the per-cell shared atom sets $S_{a,b}$
for the correlation table alongside the matrix.

\section{Cross-Axis Scoring Design}
\label{sec:decouple}

The scoring design keeps imagination-side support signals separate from
hallucination-side penalty signals whenever a task family draws on
related evidence. In PropConj, the grounding signal for $\Iassoc$ stays
distinct from the context-support check for $\Hctx$. GCW likewise
separates narrative support from unsupported-detail evidence, and CJST
separates counterfactual compliance from logic violations. For the cells
reported in Table~\maintabcorrelation{} of the \mainpaper{}, these
definitions leave $S_{a,b}{=}\varnothing$. The atom-partial correlations
therefore come out nearly identical to the unadjusted ones. The partial
analysis consequently tests formula overlap in the results we report.

\section{Decisive Cells: Full Table}
\label{app:full_cells}

\begin{table*}[t]
\centering
\includegraphics[width=\linewidth]{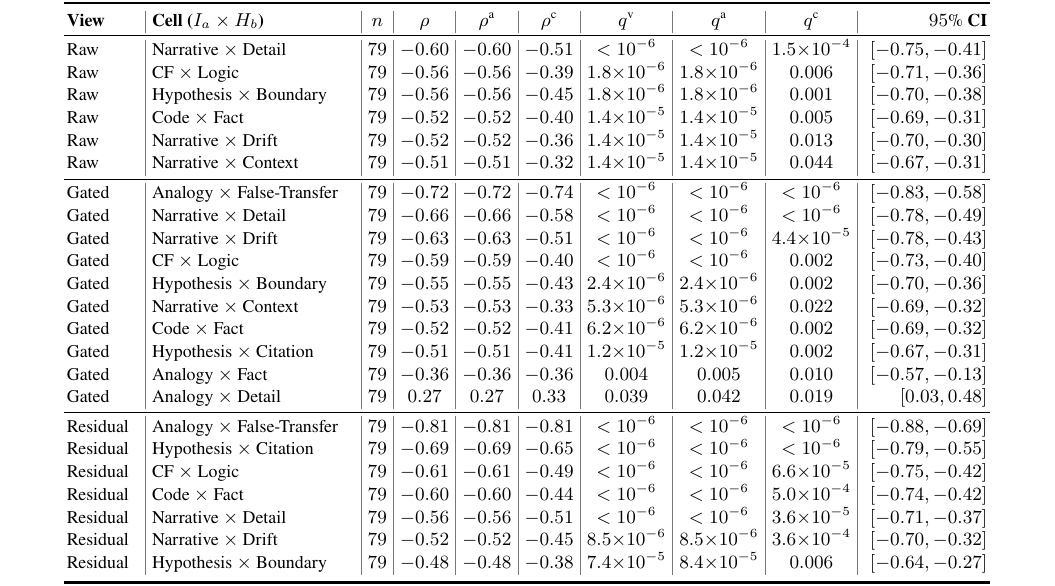}
\caption{All $23$ triple-FDR-significant cells in the main panel;
column notation follows Table~\maintabcorrelation{} of the
\mainpaper{}.}
\label{tab:full_decisive_cells}
\end{table*}

The \mainpaper{} shows only the strongest decisive cells.
Table~\ref{tab:full_decisive_cells} lists every cell clearing BH-FDR at
$q{\leq}0.05$ in its own view and in both partial controls, grouped by
view. Each row carries the raw Spearman correlation, its atom- and
capability-partial versions, the three $q$ values, and the bootstrap
interval. Twenty-three cells qualify: six in the raw view, ten in the
gated view, and seven in the residual view. Twenty-two of the
twenty-three are negative. The negative cells recur across views rather
than appearing once. Five distinct pairs qualify in all three views
(narrative $\times$ detail, narrative $\times$ drift, counterfactual
$\times$ logic, code $\times$ fact, hypothesis $\times$ boundary), and
three more qualify in two. The one positive cell qualifies in a single
view, at the largest corrected $q$ in the table.

\section{Forest Plot of Decisive Cells}
\label{app:forest}

\begin{figure*}[t]
\centering
\includegraphics[width=\linewidth]{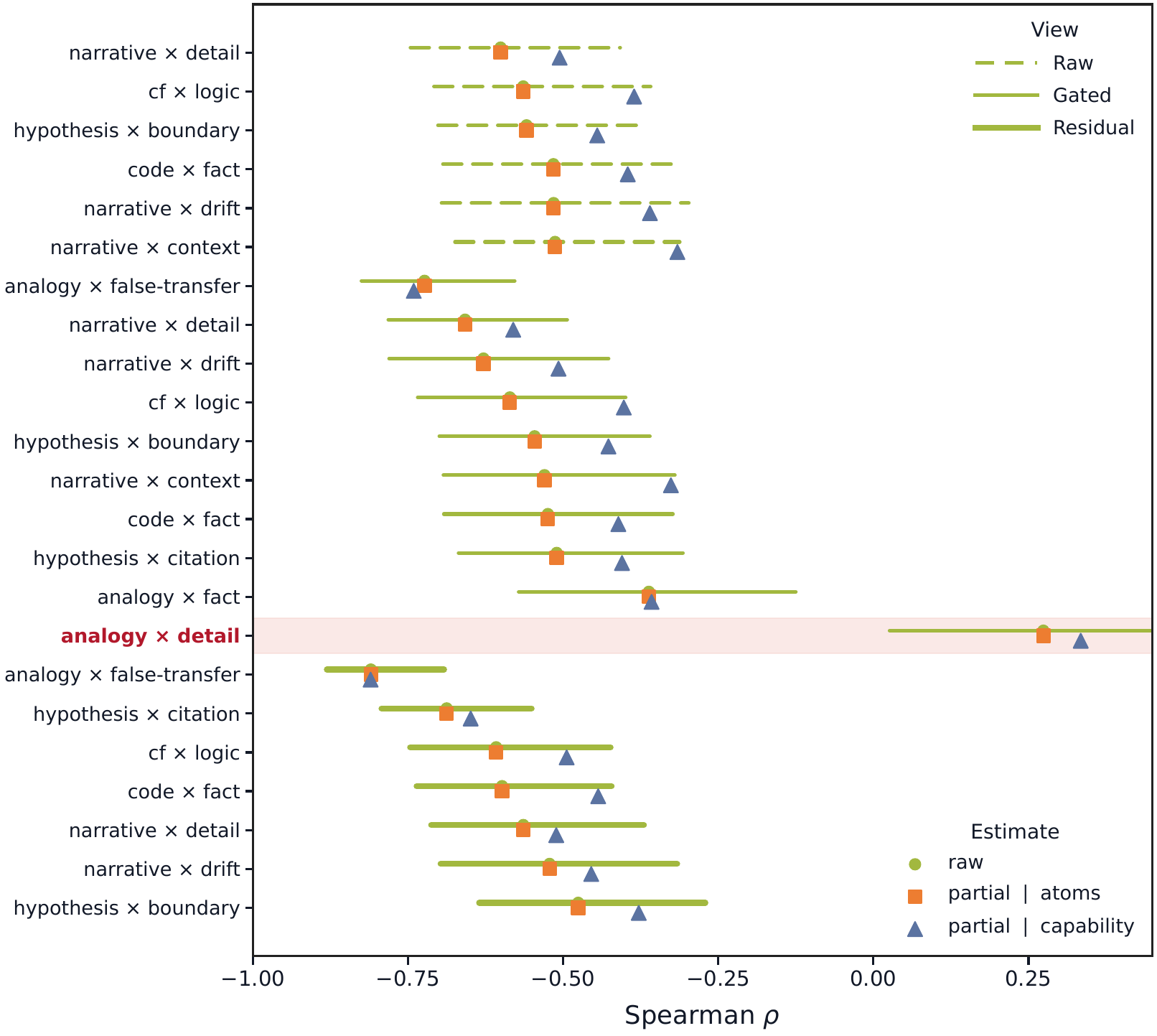}
\caption{Forest plot of the $23$ decisive cells. Raw, atom-partial, and
capability-partial estimates with $95\%$ bootstrap CIs.}
\label{fig:decisive_forest}
\end{figure*}

Figure~\ref{fig:decisive_forest} puts effect sizes and uncertainty next
to the counts above. The figure draws each decisive cell three times,
once per estimator (raw, atom-partial, capability-partial), with $95\%$
bootstrap intervals. Twenty-two intervals lie entirely below zero. One
interval, analogy $\times$ detail in the gated view, lies entirely above
zero. For almost every cell the three estimators fall within a few
hundredths of each other, so the couplings do not depend on which
control we apply. The widest intervals belong to the smallest effects,
not to the strongest ones.

\FloatBarrier
\section{Paired Discriminative Power and Reliability}
\label{app:paired}

\begin{figure}[t]
\centering
\includegraphics[width=\columnwidth]{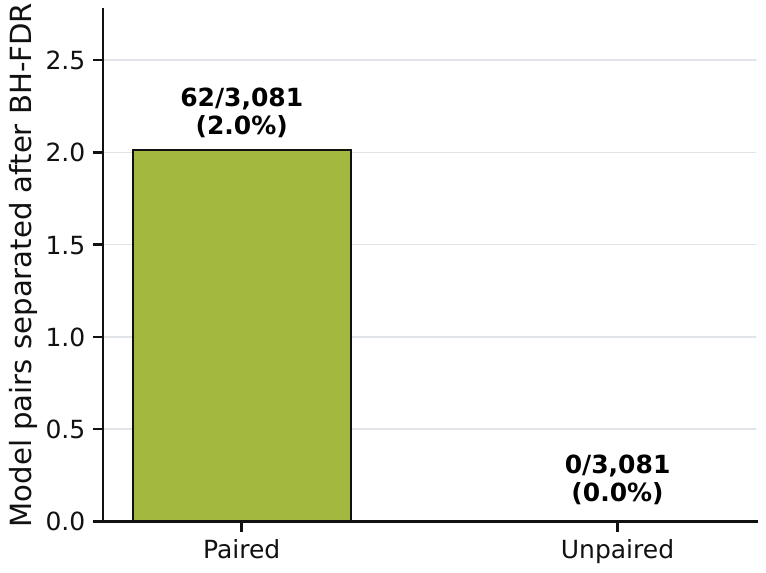}
\caption{Model pairs distinguishable after BH-FDR under paired and
unpaired analyses of the same scores.}
\label{fig:paired_power}
\end{figure}

\begin{figure*}[t]
\centering
\includegraphics[width=\linewidth]{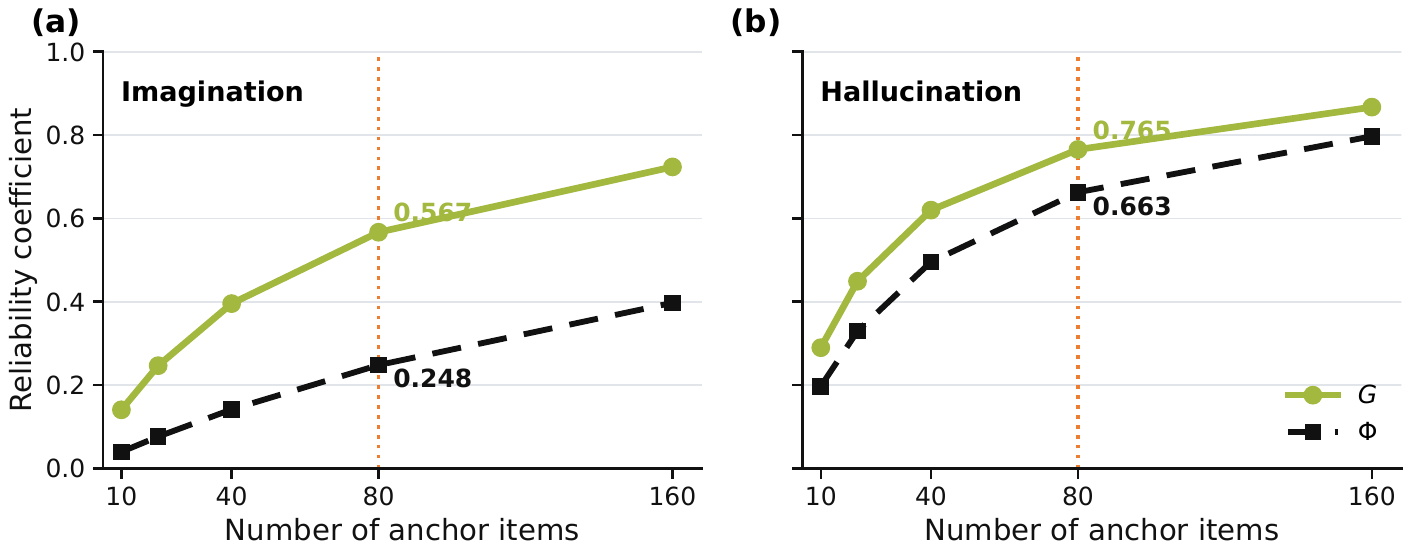}
\caption{Projected reliability as the number of anchor items changes.
Panel (a) reports imagination and panel (b) hallucination; solid green
lines show relative reliability $G$, and dashed black lines show absolute
reliability $\Phi$.}
\label{fig:reliability}
\end{figure*}

We first ask how much detection power the shared items buy.
On the $58$ models with complete coverage, we tested all $1{,}653$ model
pairs twice over the same scores: a Wilcoxon signed-rank test over
shared items, and a Mann--Whitney test without pairing. Both runs then
took BH-FDR correction.
The paired analysis separates $54$ pairs, and the unpaired analysis
separates none (Figure~\ref{fig:paired_power}). Holding the item fixed
removes prompt difficulty from the model contrast, and the removal
accounts for the gain.

How many items a comparison needs is a separate question. For the same
$58$ complete models, we decomposed model, item, and
interaction variance, then projected
relative reliability $G$ and absolute reliability $\Phi$ from $10$ to
$160$ items. At $80$ items, imagination reaches $G{=}0.567$ and
$\Phi{=}0.248$, and hallucination reaches $G{=}0.765$ and
$\Phi{=}0.663$ (Figure~\ref{fig:reliability}). Item difficulty carries
the larger share of variance on the imagination side, and the lower
absolute reliability follows. Panel-level relative comparisons therefore
rest on firmer ground than absolute score transfer to another prompt set.

\section{Sensitivity Analysis}
\label{app:sensitivity}

\begin{figure*}[t]
\centering
\includegraphics[width=\linewidth]{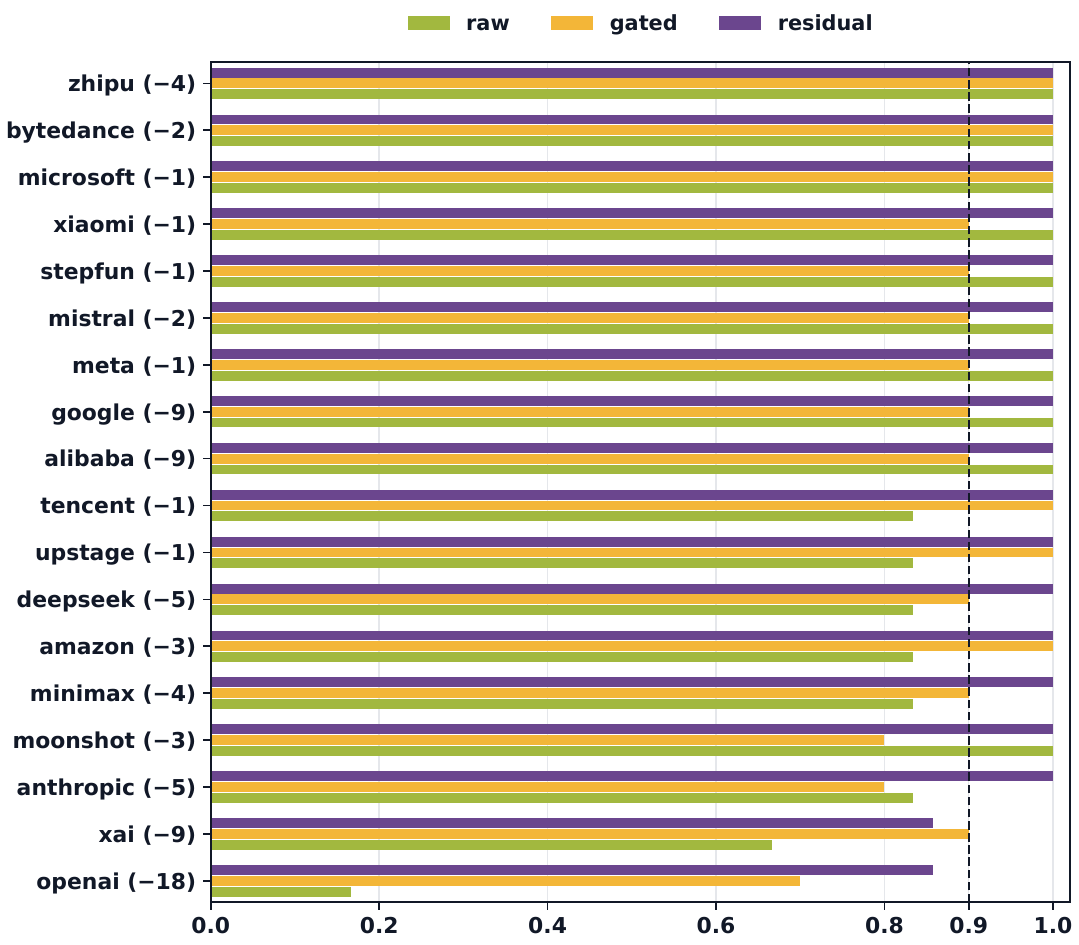}
\caption{Provider leave-one-out retention of decisive cells, by
analysis view.}
\label{fig:sensitivity_retention}
\end{figure*}

\begin{figure*}[t]
\centering
\includegraphics[width=\linewidth]{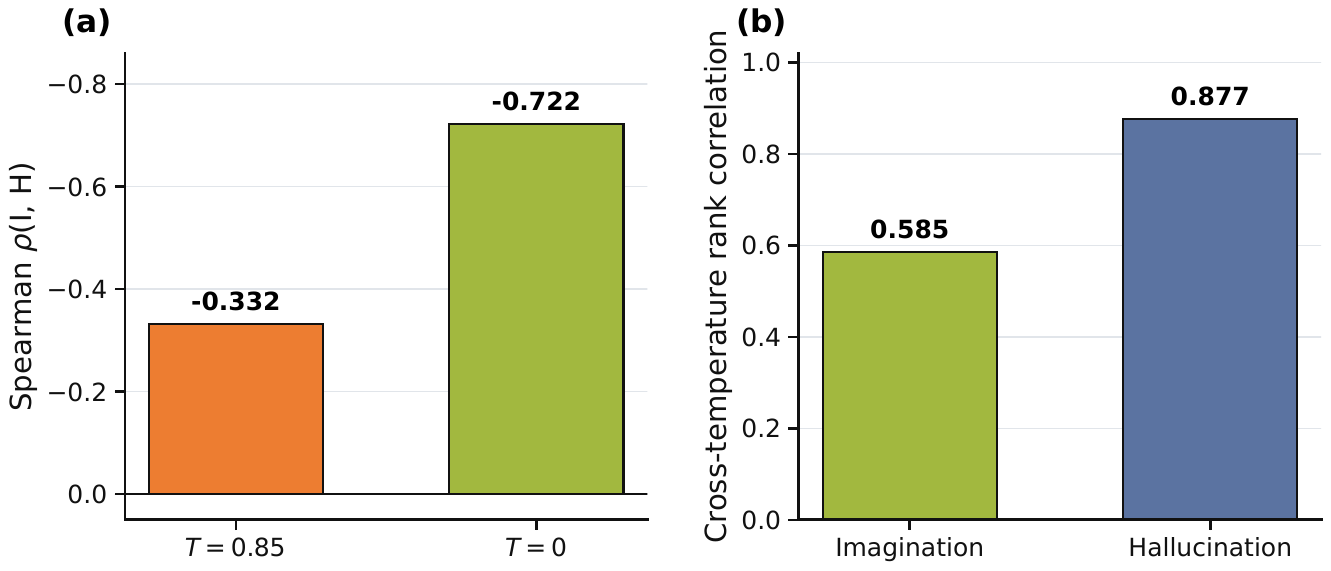}
\caption{Sensitivity to decoding temperature on $43$ matched models.
Panel (a) compares the aggregate imagination--hallucination correlation,
and panel (b) reports rank agreement between the two temperature
settings on each axis.}
\label{fig:temperature_sensitivity}
\end{figure*}

We ran two robustness checks. First, we deleted each of the $18$
providers in turn and recomputed the map. Second, we applied the same
fixed scoring rule to $43$ models generated at temperatures $0.85$ and
$0$, keeping the same $73$--$80$ valid items per model across settings.
Decisive-cell retention under provider deletion is
$17$--$100\%$ in the raw view, $70$--$100\%$ in the gated view, and
$86$--$100\%$ in the residual view. At temperature $0$, the aggregate
correlation moves from $-0.332$ to $-0.722$, with cross-temperature rank
correlations of $0.585$ for imagination and $0.877$ for hallucination
(Figures~\ref{fig:sensitivity_retention} and
\ref{fig:temperature_sensitivity}). The raw view reacts to removal
of a large provider. The gated and residual views stay more
stable, and deterministic decoding does not remove the negative
association. The gated and residual findings therefore hold across
provider composition and both decoding settings, and the raw view
carries no claim on its own.

\section{Subtype Profiles by Tier}
\label{app:radar}

\begin{figure*}[t]
\centering
\includegraphics[width=\linewidth]{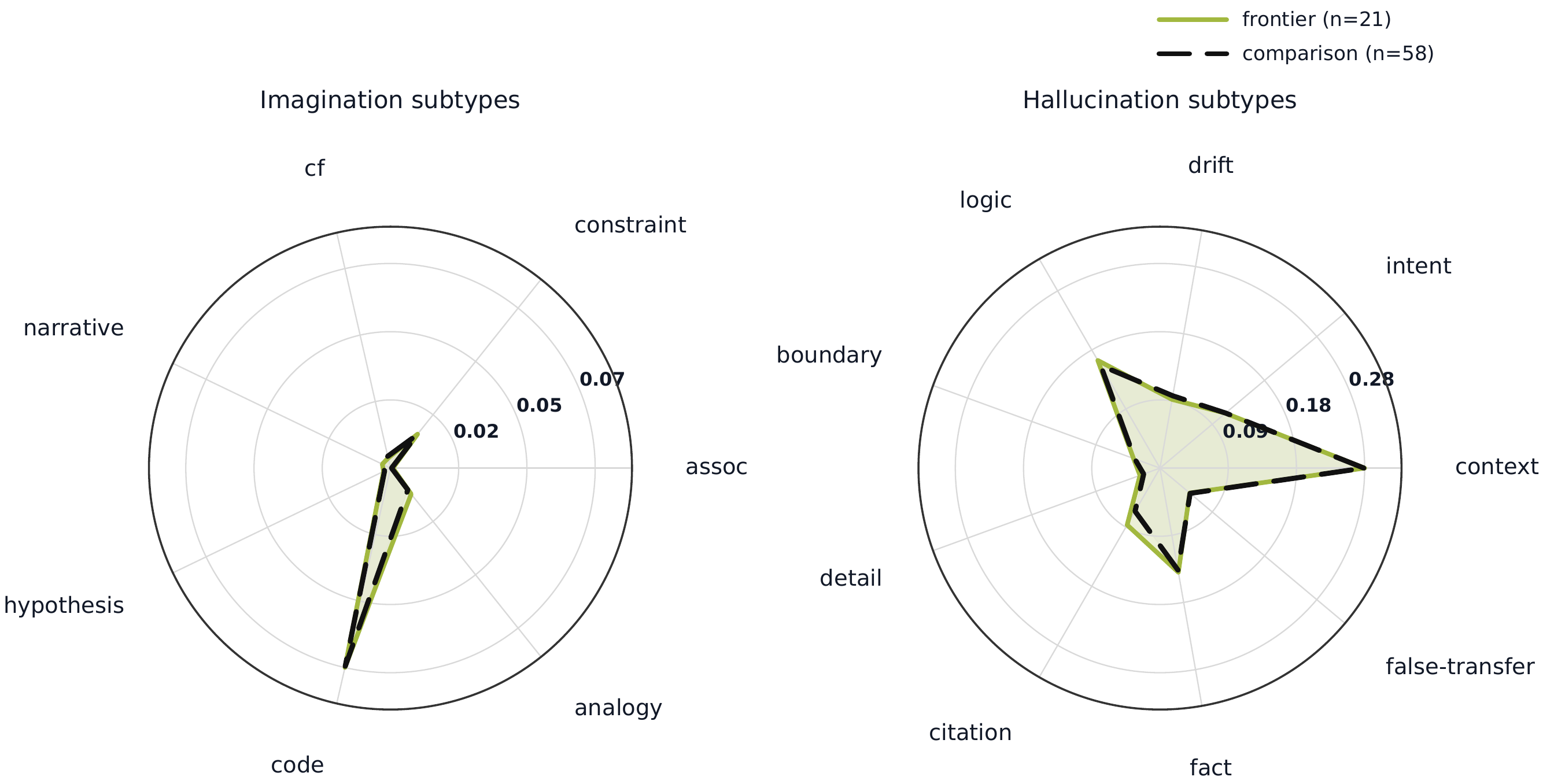}
\caption{Subtype profiles by model tier (frontier vs comparison).}
\label{fig:subtype_radar}
\end{figure*}

The \mainpaper{} reports a null tier difference on both axes, and this
section shows the profiles behind the null.
Figure~\ref{fig:subtype_radar} plots the mean score of each imagination
and destructive-hallucination subtype for the $21$ frontier and the $58$
comparison models. The two profiles overlap on both axes, and no
subtype puts one tier consistently outside the other. The tiers
differ in product position and release recency, not in the subtype
composition of either axis. The figure serves orientation only. Every
inferential claim rests on the triple-FDR map (Figure~\mainfigheatmap{}
of the \mainpaper{}) and on the tests in section
``\mainsecpurified{}''.

\section{Sensitivity Across Scoring Views}
\label{app:viewdrift}

Some cells read differently under the raw, gated, and residual scoring
views. The released tables report all three
estimates and flag every cell whose largest difference reaches
$|\Delta\rho|{\geq}0.30$. A flag alone removes nothing,
because retention still requires the view test and both partial controls
to pass the same false-discovery threshold. A large cross-view difference
instead signals that a support gate or a correction term contributes to
the observed association. Reporting all three estimates keeps a
single-view result from passing as view-invariant.

\FloatBarrier
\section{Case Study: Six Audited Outputs}
\label{app:case}

Six outputs make the joint reading concrete: for each of three task
families we pair one output whose invention stays inside the support
boundary with one whose invention crosses it. The examples are
qualitative illustrations from the benchmark corpus and carry no
statistical claim. The three families carry different imagination
subtypes; narrative $\times$ detail and cf $\times$ logic are decisive
cells of the correlation map, while the code pair is illustrative because
code $\times$ intent does not survive the triple-FDR criterion. Table~\ref{tab:case_settings} states what each prompt provides and
licenses; Tables~\ref{tab:case_study}, \ref{tab:case_cjst}, and
\ref{tab:case_code} present the three pairs.

\begin{table}[t]
\centering
\includegraphics[width=\columnwidth]{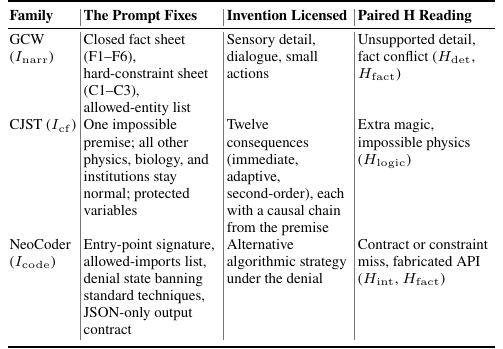}
\caption{Task settings for the three case-study families: what each
prompt fixes, what it licenses, and which hallucination subtypes read
the same output.}
\label{tab:case_settings}
\end{table}

\paragraph{Grounded Fiction (GCW).} On the card \emph{The Broken
Compass}, one output turns the floor itself into the way out:
Priya reads the raised cracks in the marble with her touch-based
left--right memory and uses them as an exit map. The turn appears
nowhere in the fact sheet, yet every element it uses does, and
\benchname{} scored it at the top of the narrative scale with no
unsupported-detail evidence. On \emph{The Fog Market}, a second
output reaches the goal only by asserting what the sheet never
supports: Otto ``recognizes the weight of the coins as those used at
Sima's spice stall,'' and Sima ``confirms the pouch belongs to a
regular customer who always buys cumin.'' Coin weight identifies
denominations, not a stall's customers, and no fact grants Sima
knowledge of the owner; \benchname{} flags both claims ($\Hdetail$,
$\Hfact$).

\begin{table}[t]
\centering
\includegraphics[width=\columnwidth]{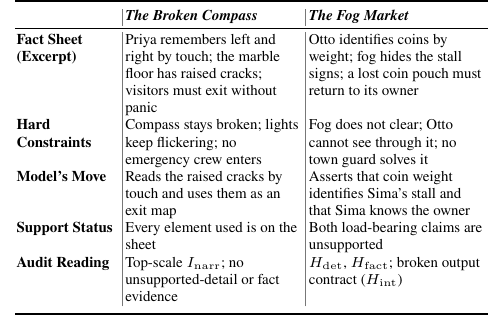}
\caption{The two GCW outputs of the case study at a
glance. The same license to invent splits into grounded invention and
fabricated shortcut exactly at the support boundary.}
\label{tab:case_study}
\end{table}

\paragraph{Counterfactual Extension (CJST).} Both outputs
were rated at the top of the counterfactual scale; only one keeps the
premise closed. Under the premise ``every cup remembers the last
drink poured into it,'' the first output derives consequences that
never leave the premise: people stop sharing cups, baristas check a
cup's memory before refilling it, parents discard cups after
allergenic drinks; each consequence carries a causal chain anchored
in the premise, and \benchname{} records no impossible-physics evidence.
Under the premise ``footsteps leave faint glowing marks for one
minute,'' the second output starts to engineer the miracle: flooring
products that ``minimize or enhance glow visibility'' and security
systems with ``one-minute glow detection algorithms.'' The premise
licenses the glow; nothing licenses machines that read or amplify it.
The prompt names this move explicitly, no extra magic beyond the
premise, and \benchname{} scored it at the top of the logic-violation
scale ($\Hlogic$).

\begin{table}[t]
\centering
\includegraphics[width=\columnwidth]{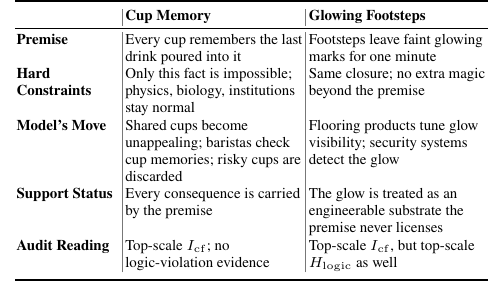}
\caption{The two CJST outputs. Both are maximally imaginative
on the imagination axis; they differ only in whether the invention stays inside
the counterfactual premise.}
\label{tab:case_cjst}
\end{table}

\paragraph{Creative Code (NeoCoder).} The third pair shares one
problem, counting 4-connected islands in a 0/1 grid, under the same
JSON-only contract: return exactly one object with the required keys,
no markdown, no text before or after it. The first output
delivers a recursive depth-first search behind the required
entry-point signature, respects the empty-imports list, and parses
cleanly; \benchname{} records no hallucination evidence on any subtype.
The second opens with reasoning text before the JSON object, so the
object never parses and the code inside never reaches the hidden
tests. \benchname{} still rated its algorithmic content at the top of
the code scale, but the delivery is exactly what $\Hintent$ records:
a contract miss that no amount of algorithmic creativity repairs.

\begin{table}[t]
\centering
\includegraphics[width=\columnwidth]{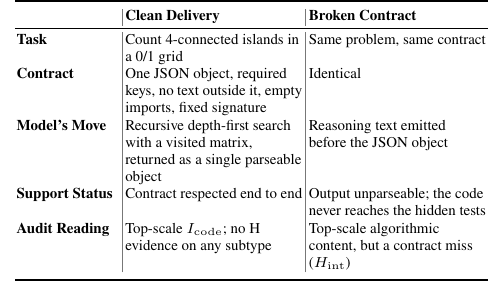}
\caption{The two NeoCoder outputs on the same count-islands
task. The pair isolates the delivery contract: identical problem,
identical license, opposite $\Hintent$ readings.}
\label{tab:case_code}
\end{table}

A distance-only creativity score cannot tell the outputs in any of
these pairs apart: within each pair both leave the obvious answer, and
in every pair both outputs were rated at the top of the imagination
scale by the scorer. The joint reading separates them, and for two of the three
families the separation it makes on single outputs is the one the
panel-scale couplings recover across $79$ models: narrative $\times$
detail at $\rho{=}{-}0.66$ and cf $\times$ logic at $-0.61$
(Table~\maintabcorrelation{} of the \mainpaper{}).

\FloatBarrier
\section{Panel Manifest}
\label{app:panel}

The full panel manifest (model identifiers, providers, families,
release dates, tier assignment, and per-item coverage) is released with
the artifact. Provider distribution of the $79$-model panel: OpenAI
($18$), xAI ($9$), Google ($9$), Alibaba ($9$), DeepSeek ($5$),
Anthropic ($5$), MiniMax ($4$), Zhipu ($4$), Moonshot ($3$), Amazon
($3$), ByteDance ($2$), Mistral ($2$), and one model each from StepFun,
Meta, Tencent, Microsoft, Upstage, and Xiaomi. The frontier tier covers the
GPT-4.1 and GPT-5 families~\citep{openai2026modelcatalog}, Claude Opus
4.6/4.7 and Sonnet 4/4.5/4.6~\citep{anthropic2026systemcards}, Gemini
3.1 Pro~\citep{google2026modelcards}, and Grok
4.x~\citep{xai2026modelcatalog}; the comparison tier adds
DeepSeek~\citep{deepseek2026changelog},
Qwen~\citep{qwen2026qwen35,qwen2026qwen36},
Kimi~\citep{moonshot2026kimik25,moonshot2026kimik26},
MiMo~\citep{xiaomi2026mimov25}, Gemma~\citep{google2026modelcards},
GLM~\citep{zai2025glm45,zai2026glm5}, Hunyuan~\citep{tencent2026hy3},
Llama~\citep{meta2024llama33},
MiniMax~\citep{minimax2026m25,minimax2026m3}, and earlier GPT, Claude,
Gemini, and ByteDance Seed
variants~\citep{seed2025thinking,bytedance2026seed2}. Tier assignment reads
model metadata only and never reads scores: $21$ models are frontier and
$58$ comparison. The panel is fully crossed by construction; the $91$
model-item cells that returned no scorable output ($79$ with no
generation, $12$ failing the task's parse contract) are recorded as
missing rather than refilled, which is why realized item overlap is
$0.98$ and not $1.00$, and why the complete-case analyses (paired
power, variance components) run on the $58$ models that carry a scored
output for every one of the $80$ items.

\section{Scoring Configuration Reproducibility}
\label{app:hyperparams}

Every downstream analysis should run off one scoring configuration. We
compared the configuration identifier attached to each of the $6{,}443$ matched annotation rows and rebuilt model
aggregates from their per-task contributions. Every matched row uses the
same configuration, and the maximum reconstruction errors are
$1.03{\times}10^{-15}$ for imagination and $4.19{\times}10^{-15}$ for
hallucination. Model identity, tier, provider, and release date are not
inputs to runtime scoring, so the reconstruction does not depend on
model-specific settings.

\paragraph{Resource Versions and Access.}
Embedding atoms use the \texttt{all-mpnet-base-v2}
encoder at revision \texttt{e8c3b32} \citep{reimers2019sbert,sentence_transformers2025allmpnet}
under the model repository's Apache 2.0 license.
Lexical resources are SWOW-EN2018 R123 \citep{dedeyne2019smallworld},
Word Norms 2 \citep{buchanan2019english}, and WordNet 3.0
\citep{miller1995wordnet}. The raw SWOW file is obtained from the official
project under its research-use terms and is not redistributed. Word Norms 2
data and its GPL-3.0 reference code are public, and the WordNet 3.0 license
permits use, copying, modification, and distribution with its notices retained.
The official Arena reference is the LM Arena method and leaderboard dataset
\citep{chiang2024chatbotarena,lmarena2026leaderboard}, using the 2026-04-27
snapshot. The frozen configuration records the final-panel Seed 1.6 anchor
as a value from the third-party BenchLM model page \citep{benchlm2026seed16},
and not an official LMArena row. The current page exposes no sourced score,
so the citation documents that limitation; it does not validate the value.
The common-answer banks are author-built: the static bank and curated overlay
are included with the artifact; mined extensions
require regeneration from the documented settings and access to the listed model APIs.

\section{Held-Out Annotator Agreement}
\label{app:calibration_agreement}

\begin{figure*}[t]
\centering
\includegraphics[width=\linewidth]{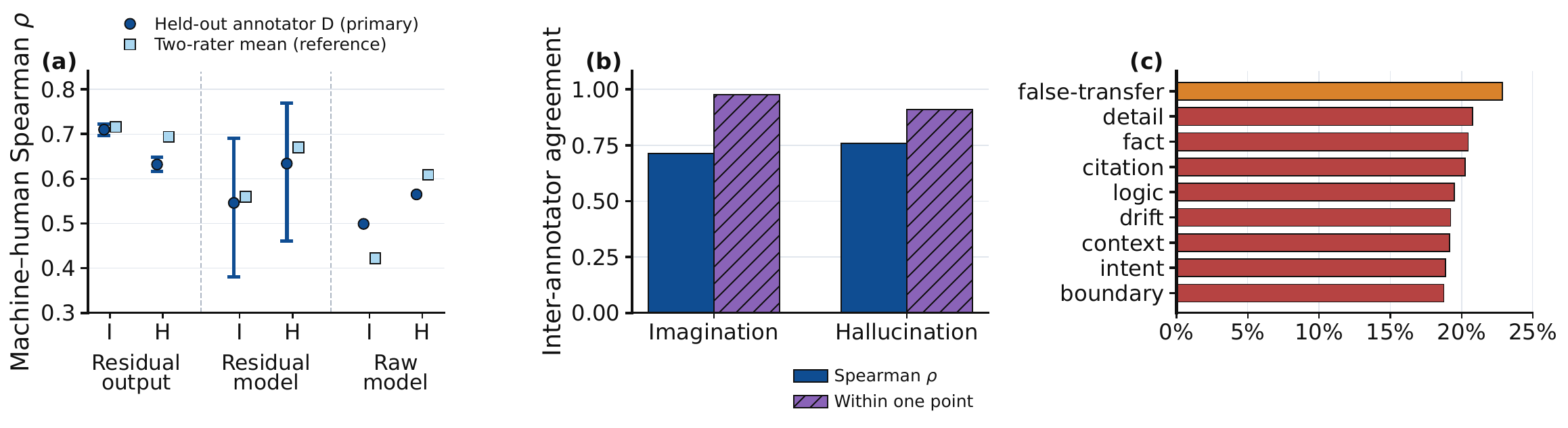}
\caption{Human-rating agreement. Panel (a) compares the residual scores
with the held-out annotator and the two-rater mean at output and model
levels; the raw model-level points provide an uncalibrated reference.
Filled dark circles mark the held-out comparison, and light squares
mark the two-rater mean. Error bars are $95\%$ bootstrap intervals for
the held-out comparisons. Panel (b) reports inter-annotator
agreement, and panel (c) shows the licensed share of unsupported content
by subtype.}
\label{fig:calibration_agreement}
\end{figure*}

Two annotators independently rated $6{,}640$ outputs on $0$--$4$
imagination and unsupported-content scales. The ratings of annotator~C
selected the subtype score boundaries; the ratings of annotator~D were
never read by any fitting step. Machine correlations with the held-out annotator are $0.710$
and $0.632$ at output level ($n{=}6{,}443$) and $0.546$ and $0.634$ at
model level ($n{=}79$). Against the two raters' mean, the corresponding
values are $0.716$ and $0.694$, and $0.560$ and $0.670$. Reversing the
split and selecting boundaries on annotator D gives model-level
correlations of $0.511$ and $0.644$ on annotator C, against $0.546$
and $0.634$ in the deployed direction. The two boundary sets do not
match cell by cell, but their model-score rankings agree at $\rho{=}0.977$
and $0.996$ for imagination and hallucination. The raw view, which the
boundary and residual calibration does not touch, agrees with the
held-out annotator at $\rho{=}0.50$ and $0.57$ at model level
(Figure~\ref{fig:calibration_agreement}a). Inter-annotator correlations
are $0.712$ for imagination and $0.759$ for unsupported content, with
agreement within one point at $97.70\%$ and $91.07\%$. Licensing rates
span $18.74\%$--$22.85\%$ across subtypes and false transfer does not
differ from the others (Fisher $p{=}.218$). The hold-out is at the rater
level. The held-out annotator worked from the same rubric on the same
outputs, so this design controls for rater-specific fitting, not for
circularity introduced by the rubric itself.

\section{Prompt Examples}
\label{app:prompt_examples}

\begin{figure*}[t]
\centering
\includegraphics[width=\linewidth]{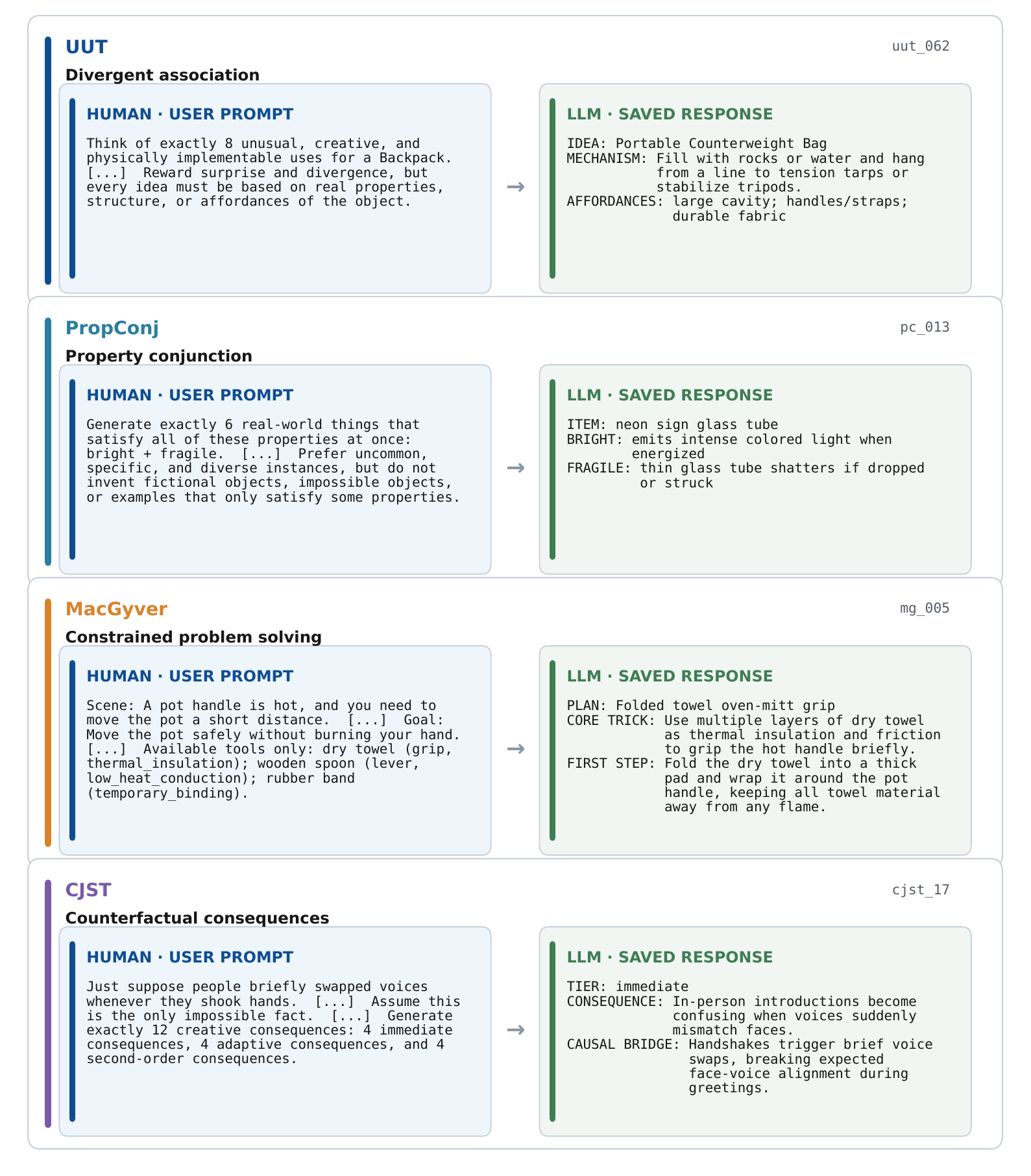}
\caption{Human--LLM examples for UUT, PropConj, MacGyver, and CJST. Each
row places evaluated user-prompt passages on the left and selected field
values from the corresponding saved response on the right. Bracketed
ellipses mark omitted prompt text, and task IDs identify the full saved
records. Colors distinguish benchmark components and do not encode
scores.}
\label{fig:prompt_examples}
\end{figure*}

\begin{figure*}[t]
\centering
\includegraphics[width=\linewidth]{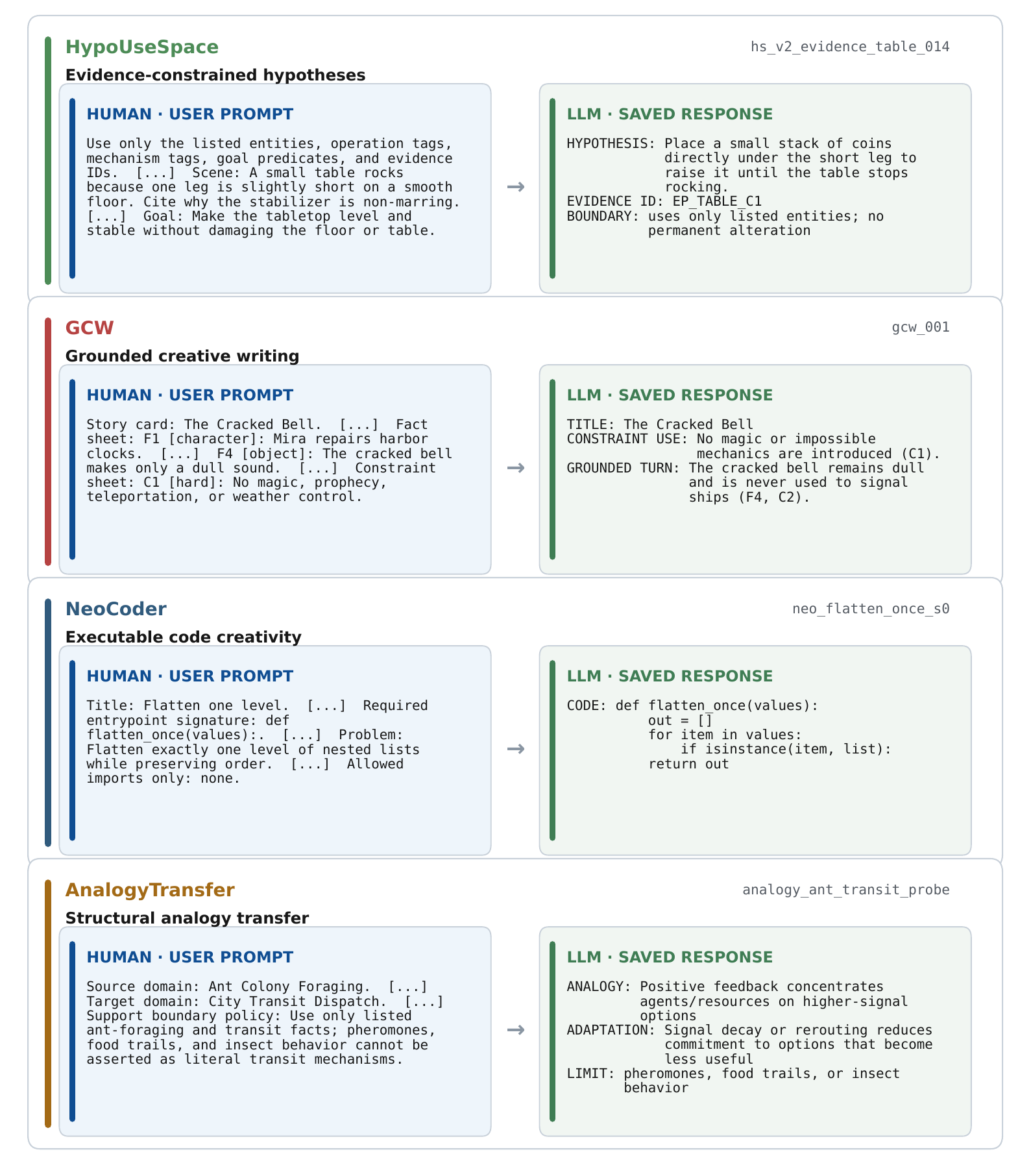}
\caption{Human--LLM examples for HypoUseSpace, GCW, NeoCoder, and
AnalogyTransfer. The left column reproduces passages from the evaluated
user prompts; the right column reports selected field values from the
saved responses. Bracketed ellipses mark omitted prompt text. Task IDs
identify the full saved records, and colors carry no score information.}
\label{fig:prompt_examples_grounded}
\end{figure*}

The crossed panel used $80$ anchor prompts from eight scored components.
We sent the same $80$ task IDs to all $79$ models. Each request contained
a task-specific system message and one user message, and asked for a
single model response. Figures~\ref{fig:prompt_examples} and
\ref{fig:prompt_examples_grounded} pair prompt passages with selected
fields from the saved responses for all eight components.

\subsection{Request Format}

The system message defined the benchmark role and output format. The
user message supplied the task instance and its support rule, followed
by the required JSON schema and output count. Depending on the task,
the instance could contain a tool inventory, a fact sheet, an evidence
table, or a function signature. Responses had to contain JSON without
surrounding prose. We did not add demonstrations or follow-up turns for
particular models.

The support rule was therefore part of the input seen by the model.
Its form varied by task. Some prompts limited proposals to physical
properties or listed tools, while others licensed a single
counterfactual premise, a fictional world, or an analogy. The prompt
also stated which additions fell outside that license.

\subsection{Task-Specific Instructions}

\paragraph{UUT and PropConj.}
UUT requested eight unusual but physically implementable uses of an
ordinary object and ruled out invented capabilities. PropConj requested
six real objects that satisfied every named property. An uncommon answer
was acceptable only when it remained possible and met the full
conjunction.

\paragraph{MacGyver and HypoUseSpace.}
MacGyver supplied a goal, a tool inventory, and physical or safety
constraints. Plans could combine listed tools but could not introduce
new equipment. HypoUseSpace instead supplied entities, relations,
predicates, and evidence IDs. Its answers had to cite the records used
to support the proposed hypothesis.

\paragraph{CJST and GCW.}
CJST introduced one impossible premise and requested immediate,
adaptive, and second-order consequences while keeping ordinary
constraints in force. GCW paired a story card with a fact sheet and a
list of forbidden claims. Narrative detail was open, but the stated
world facts and exclusions remained binding.

\paragraph{NeoCoder and AnalogyTransfer.}
NeoCoder fixed the entry point, required behavior, permitted imports,
and prohibited techniques for an executable coding task.
AnalogyTransfer named a source system and a target domain, then specified
which source relations could transfer and which source details could not
be treated as literal facts about the target.

\subsection{Source of the Examples}

The cards draw from the prompt manifest and task results saved with one
panel report. The build script checks each displayed prompt passage
against the full prompt and each response field value against the raw
model output; it stops if either check fails. A bracketed ellipsis marks
text omitted between prompt passages. Each task ID locates the complete
prompt and response. Colors distinguish components and do not encode
scores.

\FloatBarrier
\section{Limitations}
\label{app:limitations}

\paragraph{Correlation, Not Causation.}
Triple controls and per-item replication identify co-variation, not
directional generative mechanism. A causal account
would require interventional fine-tuning that selectively reduces one
subtype, which we leave to follow-up work. The same controls reduce but
cannot fully eliminate formula-driven correlation, since some atoms
straddle the I/H boundary; cells whose estimate shifts across views are
flagged in Appendix~\ref{app:viewdrift}.

\paragraph{The Machine--Human Coupling Gap.}
The two axes track human ratings individually, but the human ratings do
not independently reproduce the negative coupling between them
($-0.14$, $95\%$ CI $[{-}0.37,{+}0.11]$, against $-0.41$ on the machine
side, both reported in the \mainpaper{}). A single $0$--$4$ rating per
axis may simply lack the resolution of sixteen subtype channels, and the
human interval does not exclude a moderate negative coupling, but on
present evidence the coupling is a property of \benchname{}'s scores
that human raters have not confirmed at strength.

\paragraph{Measurement Resolution.}
Items carry $74.5\%$ of the score variance and models $0.41\%$, giving
$G{=}0.567$ and $\Phi{=}0.248$ at $80$ items and leaving only $3.3\%$ of
model pairs separable. Absolute scores are therefore not comparable
across item sets, and individual model rankings should not be read off
these numbers. Five of the sixteen subtype channels also place more than
half their outputs at the floor, which compresses the correlations those
channels can express.

\paragraph{Scope and Resource Dependence.}
The map is only as general as the panel; generalization beyond
$2026$-vintage instruction-tuned LLMs is unclaimed. \benchname{} is
deterministic and auditable but not resource-free: it depends on
SBERT-class embeddings, association and feature norms, WordNet, curated
common-answer banks, closed-world manifests, and hidden tests, so we
name the scoring \emph{deterministic and embedding-anchored} rather than
symbolic. Proxy swaps of the embedding model and of bank coverage, run
on the earlier prompt-collection panel, retained only a minority of
decisive cells, so these resources carry structural weight on the
imagination side, and extension to another language requires rebuilding
them rather than translating prompts. Deterministic atoms are also blind
to higher-order qualities such as narrative tension, which a rubric-based
judge would see at the cost of auditability.

\section{Statistical Power and Design Sensitivity}
\label{app:power_analysis}

\begin{figure*}[t]
\centering
\includegraphics[width=\linewidth]{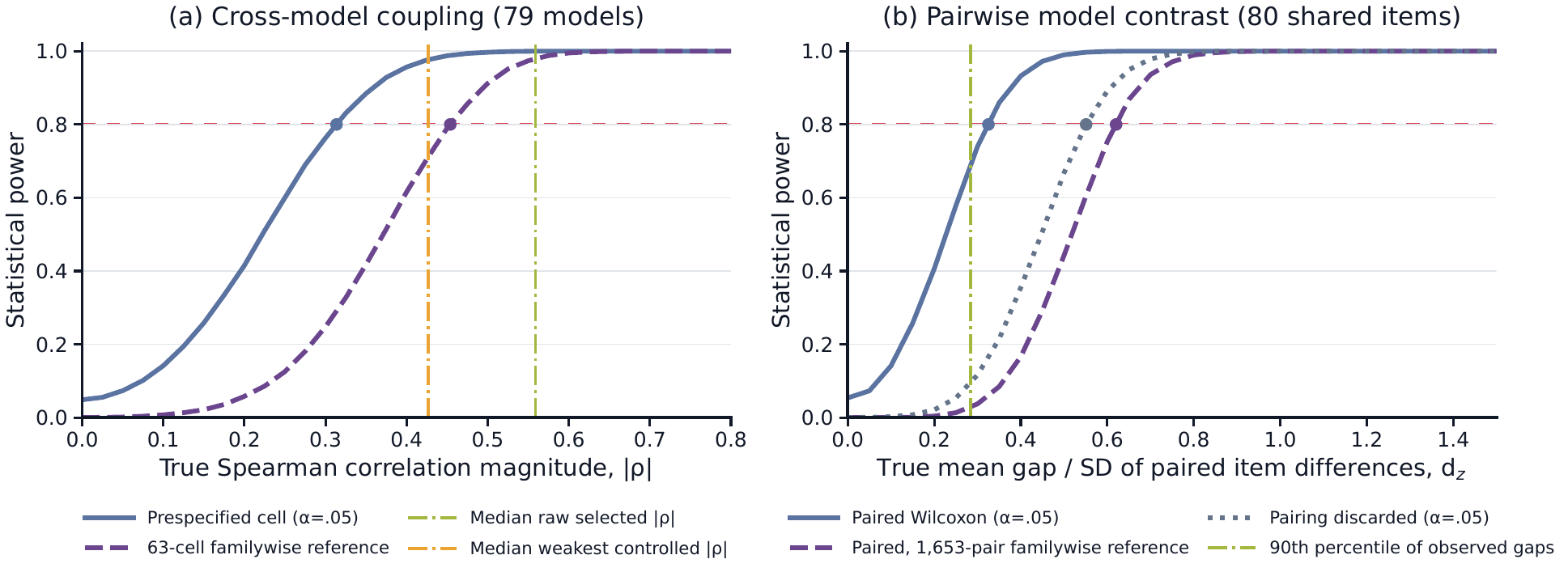}
\caption{Statistical power under the current benchmark design. Panel
(a) reports Monte Carlo power for a two-sided Spearman correlation over
$79$ models; the dashed curve uses $\alpha{=}.05/63$ as a conservative
familywise reference for one $63$-cell view, and the vertical lines mark
the median raw and weakest controlled coefficients among the selected
cells. Panel (b) reports variance-calibrated power for an
imagination-score difference over $80$ shared items; the curves compare
the paired Wilcoxon test at $\alpha{=}.05$, a conservative
$\alpha{=}.05/1653$ reference, and the Mann--Whitney $U$ test after
discarding item pairing. Points mark the minimum effect reaching $80\%$
power, and the vertical line marks the $90$th percentile of observed
model-mean gaps.}
\label{fig:statistical_power}
\end{figure*}

What effect sizes can $79$ models and $80$ shared items actually resolve?
Because data collection was complete, we estimated design sensitivity
rather than achieved power, in $10{,}000$ Monte Carlo runs (seed $20260729$),
applying Spearman tests to Gaussian-copula samples of $79$ models and
paired Wilcoxon tests to variance-calibrated Gaussian random-effects
samples for $58$ complete models and $80$ shared items.
At nominal $\alpha{=}.05$, $80\%$ power was reached at
$|\rho|{=}.31$ for one coupling and an imagination-score gap of
$.041$ ($d_z{=}.33$); conservative familywise references raised the
thresholds to $|\rho|{=}.45$ for $63$ cells and $.078$
($d_z{=}.62$) for $1{,}653$ pairs
(Figure~\ref{fig:statistical_power}a--b). The selected cells have median
raw $|\rho|{=}.56$, above the conservative correlation threshold,
and shared-item pairing reduced the detectable imagination gap by
$41\%$ relative to the $.069$ unpaired threshold, a gain consistent
with items carrying $74.5\%$ of the score variance. The current design
therefore has adequate sensitivity for the moderate-to-large couplings
that support the main finding, while the shared anchor set improves
model contrasts, although small controlled associations and closely
spaced model pairs remain below its resolution.

\FloatBarrier
\section{Supplementary Related Work}
\label{app:single_axis_benchmarks}

The main paper focuses on work that measures imagination and
hallucination together. This section expands that discussion by
reviewing the two single-axis benchmark traditions that motivate
\benchname{}'s paired design.

\subsection{Creativity Benchmarks for LLMs}
\label{app:rel_creativity}

Creativity benchmarks operationalize divergent production through
several task-specific measures. The Divergent Association Task scores
semantic distance among words generated to be mutually
unrelated~\citep{olson2021naming}. Automated scoring of the Alternative
Uses Test instead measures semantic distance between an alternative use
and the object named in the prompt~\citep{beaty2021forward}. Forward
flow measures how far each word in a free-association chain moves from
earlier words~\citep{gray2019forward}. \citet{stevenson2022putting} apply the
Alternative Uses Test to GPT-3 and compare its outputs with human
responses using expert ratings of originality, usefulness, surprise,
and flexibility, together with automated semantic-distance
scores. CreativityPrism groups tasks from
divergent thinking, creative writing, and logical reasoning under
quality, novelty, and diversity, using automatic judges validated
against human annotations~\citep{hou2025creativityprism}.
LiveIdeaBench elicits scientific ideas from single-keyword prompts and
evaluates them with a dynamic multi-model judging
panel~\citep{ruan2024liveideabench}. MacGyver tests object reuse under
physical constraints~\citep{tian2024macgyver}, while CS4 varies the
number of story-writing constraints to assess creativity without human
ratings~\citep{atmakuru2024cs4}. NeoCoder uses denial prompting to test
divergent and convergent thinking on programming
problems~\citep{lu2024neocoder}. \citet{alrabeyah2025do} test whether four LLM
judges agree with an oracle set of Alternative Uses Test responses and
with one another. These benchmarks measure
creativity or its evaluation, but they do not assign paired imagination
and support-boundary labels to the same generation. \benchname{} adds
that paired reading while retaining task-specific measures of divergent
production.

\subsection{Hallucination Benchmarks for LLMs}
\label{app:rel_hallucination}

Hallucination benchmarks range from answer-level truthfulness tests to
fine-grained support checks. TruthfulQA tests whether models reproduce
common human misconceptions in answers to adversarially selected
questions, using generation and multiple-choice
scores~\citep{lin2022truthfulqa}. HaluEval provides generated and
human-annotated hallucination examples for question answering,
dialogue, and summarization~\citep{li2023halueval}. FActScore decomposes
long-form generations into atomic facts and verifies each against a
knowledge source~\citep{min2023factscore}. LongFact supplies open-domain
prompts, while SAFE decomposes responses and checks individual facts
with search~\citep{wei2024longfact}. ALCE evaluates retrieval-augmented
answers using citation correctness and completeness, alongside fluency
and correctness~\citep{gao2023alce}. RAGTruth annotates hallucinations at
both response and word levels in retrieval-grounded question answering
and summarization~\citep{niu2024ragtruth}. HALoGEN verifies atomic units
across nine domains and distinguishes errors associated with incorrect
recall, source knowledge, and fabrication
~\citep{ravichander2025halogen}. HalluLens separates intrinsic and
extrinsic hallucination tasks~\citep{bang2025hallulens}, while
\citet{huang2024hallusurvey} organize definitions, causes, detection,
and mitigation methods. Across these benchmarks, the unit
of evaluation becomes progressively finer, but the labels remain on the
truthfulness or support side. The cited benchmarks do not also assign an
imagination subtype to the same output. \benchname{} pairs that
support-boundary reading with ten hallucination subtypes and seven
imagination subtypes on each generation.

\section{Reliability and the Boundaries of the Instrument}
\label{app:instrument_reliability}

We close with the resolution of the instrument itself.
Among the $58$ complete models, items
account for $74.5\%$ of variance and models for $0.41\%$; reliability is
$G{=}0.567$ and $\Phi{=}0.248$. Five subtypes are compressed at the floor,
while provider deletion preserves at least $70\%$ of decisive cells in
the gated and residual views (Appendix~\ref{app:sensitivity}). With only
$3.3\%$ of model pairs separable (section ``\mainsecanchor{}'' of the
\mainpaper{}), the instrument supports panel-level structure rather than
individual rankings. Further boundaries are detailed in
Appendix~\ref{app:limitations}.

\FloatBarrier
\end{document}